\documentclass[a4paper,fleqn]{cas-dc}

\usepackage{graphicx}
\usepackage{amsmath}
\usepackage{amssymb}
\usepackage{array}
\usepackage{booktabs}
\usepackage{multirow}
\usepackage[numbers]{natbib}

\ExplSyntaxOn
\cs_set:Npn \__first_footerline:
  {
    \group_begin:
    \footnotesize\rmfamily
    \href{https://doi.org/10.1016/j.array.2026.101421}
      {doi:10.1016/j.array.2026.101421}
    \quad
    \href{https://creativecommons.org/licenses/by/4.0/}{CC~BY~4.0}
    \group_end:
  }
\ExplSyntaxOff

\begin{document}

\let\WriteBookmarks\relax
\def\floatpagepagefraction{1}
\def\textpagefraction{.001}

\shorttitle{Temporal pseudo-RGB encoding for label-free VEEV detection}
\shortauthors{Z. Zhu et al.}

\title[mode=title]{Temporal Pseudo-RGB Encoding for Label-Free Detection of VEEV Infection in Human Brain Endothelial Cells}

\author[1]{Zhengjie Zhu}[orcid=0009-0000-0251-5786]
\cormark[1]
\ead{Zhengjie.Zhu@wfusm.edu}
\credit{Conceptualization, Methodology, Software, Validation, Formal analysis, Investigation, Data curation, Writing -- original draft, Writing -- review and editing, Visualization}

\author[2,3]{Stephanie Valdez Trefry}[orcid=0000-0002-2039-1825]
\credit{Validation, Investigation, Data curation, Writing -- review and editing}

\author[1]{Hao Lu}[orcid=0000-0002-4547-7780]
\credit{Writing -- review and editing}

\author[1]{Abbas Alili}[orcid=0000-0002-7849-6881]
\credit{Writing -- review and editing}

\author[1]{Yongxin Guo}[orcid=0009-0002-8377-0299]
\credit{Writing -- review and editing}

\author[2]{Aarthi Narayanan}[orcid=0000-0002-4889-4416]
\credit{Conceptualization, Resources, Supervision, Writing -- review and editing}

\author[1]{Metin Nafi Gurcan}[orcid=0000-0002-2421-8229]
\credit{Conceptualization, Resources, Supervision, Project administration, Writing -- review and editing}

\affiliation[1]{organization={Wake Forest University School of Medicine},
    addressline={Center for Artificial Intelligence Research},
    city={Winston-Salem},
    state={NC},
    postcode={27157},
    country={USA}}

\affiliation[2]{organization={George Mason University},
    addressline={Department of Biology},
    city={Fairfax},
    state={VA},
    postcode={22030},
    country={USA}}

\affiliation[3]{organization={George Mason University},
    addressline={School of Systems Biology},
    city={Manassas},
    state={VA},
    postcode={20110},
    country={USA}}

\cortext[1]{Corresponding author}

\begin{abstract}
Label-free detection of viral infection is central to antiviral screening and pathogen surveillance. Most deep-learning methods analyze individual frames and do not model temporal progression. We evaluate temporal pseudo-RGB encoding, which stacks three time-offset brightfield frames ($t{-}6$\,h, $t{-}3$\,h, $t$) as the red, green, and blue channels of a single image so that morphological change becomes color and an ImageNet-pretrained convolutional neural network (CNN) can be trained on infection trajectories without architectural modification. In human brain microvascular endothelial cells (HBMVEC) infected with Venezuelan equine encephalitis virus (VEEV) TC-83, a multi-task network predicts infected-versus-mock status and time post-infection in one forward pass. Across three experimental runs under leave-one-row-out cross-validation, with all 36 wells scored out of fold, the encoding reaches 91.6\% accuracy against 87.4\% for the single-frame model (paired difference $+4.2$ percentage points, 95\% CI $[+2.5, +6.1]$; sign-flip $p = 3\times10^{-5}$; well-level inference, conditional on three runs). At the late-phase endpoint (40--48\,h), where floating cells and debris can make conventional brightfield images difficult to interpret, the advantage widens to $+23.9$ points (95\% CI $[+19.2, +28.9]$). Time post-infection is recovered on the fixed row split with MAE = 1.31\,h ($R^2 = 0.975$). Among the temporal models tested the encoding has the highest mean accuracy, at a fraction of the cost of the recurrent and 3D convolutional alternatives, though only its margin over the single-frame model is resolved; the tested ViT-B/16 backbone showed no corresponding gain, so the finding is limited to convolutional backbones and this virus--cell system.

\end{abstract}

\begin{keywords}
Deep learning \sep Temporal encoding \sep Label-free microscopy \sep Viral infection \sep Multi-task learning \sep Infection detection \sep Computer vision
\end{keywords}

\maketitle

\hypersetup{
  pdftitle={Temporal Pseudo-RGB Encoding for Label-Free Detection of VEEV Infection in Human Brain Endothelial Cells},
  pdfauthor={Zhengjie Zhu, Stephanie Valdez Trefry, Hao Lu, Abbas Alili, Yongxin Guo, Aarthi Narayanan, Metin Nafi Gurcan},
  pdfsubject={Label-free detection of viral infection in time-lapse brightfield microscopy},
  pdfkeywords={Deep learning; Temporal encoding; Label-free microscopy; Viral infection; Multi-task learning; Infection detection; Computer vision}
}

\section{Introduction}
Automated detection of viral infection from label-free microscopy could support high-throughput antiviral drug screening, vaccine and neutralizing-antibody evaluation, biodefense and emerging-pathogen surveillance, and basic studies of viral kinetics \cite{ref-morens2004emerging,ref-woolhouse2005emerging,ref-leland2007cpe}.
Brightfield imaging is well suited to these applications: it is non-invasive, compatible with continuous longitudinal imaging at high temporal resolution, and free of the artifacts and endpoint constraints associated with fluorescent labels or destructive viability assays \cite{ref-riss2016assay,ref-petkidis2023review}.
Here, \emph{label-free} refers to the imaging: the images are brightfield frames of live, unfixed cultures, and no stain or fluorescent label is imaged or used to derive the training labels.
The models are trained with supervision taken from the experimental design, namely the infection condition of each well and the time elapsed since inoculation, so no annotation of individual cells is required.
Recent deep-learning methods detect cytopathic effects (CPE)---the cell rounding, shrinkage, and detachment that mark productive viral infection---directly from brightfield images \cite{ref-andriasyan2021iscience,ref-petkidis2024natcomm,ref-chen2022biomedicines,ref-airvic2025}.
All these methods, however, treat each frame independently and discard the temporal trajectory information inherent in time-lapse imaging.

Because infection unfolds over time, a model trained on isolated snapshots must infer progression from a single static appearance.
This limitation is likely to be most important when the current appearance is ambiguous, such as during very early stages before overt CPE emerges or during late stages when floating cells and debris can obscure monolayer morphology in a single brightfield frame \cite{ref-wang2020cpe,ref-hsiung1994cpe,ref-leland2007cpe}.
The importance of analyzing temporal change rather than static appearance has long been recognized in computer-aided medical image analysis, particularly in longitudinal disease assessment \cite{ref-hadjiiski2001temporal,ref-prescott2010temporal}.
We reasoned that encoding the \emph{trajectory} of morphological change, rather than the instantaneous state, could make label-free detection more stable across the infection timeline.

\begin{figure*}[t]
  \centering
  \includegraphics[width=\linewidth]{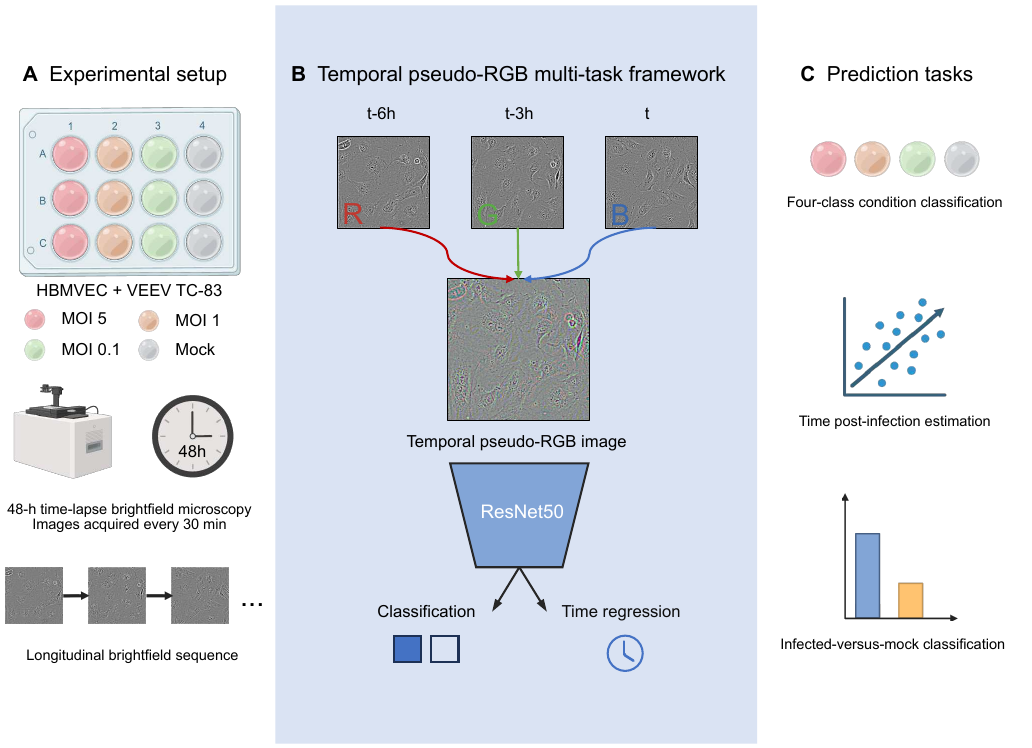}
  \caption{End-to-end overview of temporal pseudo-RGB encoding. HBMVEC cultures infected with VEEV TC-83 at MOI~5, MOI~1, MOI~0.1, or mock are imaged in brightfield every 30 minutes over 48 hours. Three time-offset frames ($t{-}6$\,h, $t{-}3$\,h, $t$) become the red, green, and blue channels of one image, so that morphological change appears as color and static regions remain gray. An ImageNet-pretrained ResNet50 with a four-class condition head and a continuous time post-infection regression head is trained on that image; the binary infected-versus-mock readout is derived by collapsing the condition predictions. Panel C names the three prediction tasks; all quantitative results are in the tables and in Figures~\ref{fig:4} and~\ref{fig:2}.}
  \label{fig:1}
\end{figure*}

To address this, we adopt temporal pseudo-RGB encoding: three time-offset brightfield frames ($t{-}6$\,h, $t{-}3$\,h, $t$) are stacked as the red, green, and blue channels of a single temporal pseudo-RGB image, so that regions undergoing morphological change appear as color shifts while static regions appear gray.
Temporal information has been incorporated into imaging models through recurrent networks, 3D convolutions, optical flow, frame differencing, and longitudinal feature analysis \cite{ref-donahue2015lrcn,ref-tran2015c3d,ref-simonyan2014twostream,ref-lee2018mfnet,ref-hadjiiski2001temporal,ref-prescott2010temporal}, and the specific device of writing time-offset frames into the color channels of a 2D network is itself established in computer vision.
Section~\ref{sec:related} reviews those methodological precedents and positions the present study relative to them.
The representation allows a standard ImageNet-pretrained CNN to repurpose its learned color-processing features for temporal discrimination, without architectural modification, additional parameters, or purpose-built temporal modules such as 3D convolutions, recurrent networks, or optical-flow estimators.
Combined with a unified multi-task training objective that jointly optimizes four-class condition classification and continuous time post-infection regression, the same forward pass yields a stable binary infected-versus-mock readout, obtained by collapsing the condition predictions, and a continuous time post-infection estimate throughout 48 hours of imaging (Figure~\ref{fig:1}).

We evaluated this approach on human brain microvascular endothelial cells (HBMVEC), which constitute a critical component of the blood--brain barrier (BBB) \cite{ref-daneman2015bbb,ref-abbott2010bbb}, infected with Venezuelan equine encephalitis virus (VEEV) TC-83, an attenuated vaccine strain of this positive-sense single-stranded RNA alphavirus.
We used TC-83 as a model with rapid replication and overt CPE to develop this methodology. Pathogenic VEEV strains are classified as Category B select agents and can traverse the BBB through infection of brain endothelial cells \cite{ref-weaver2004veev,ref-schafer2011vee,ref-salimi2020mbio}.
This setting provides a biologically relevant test bed for label-free monitoring of infection dynamics in time-lapse brightfield microscopy.

This work makes four contributions.
First, we characterize temporal pseudo-RGB encoding in a regime in which it has not previously been tested: offsets of hours rather than milliseconds, and a slow, non-rigid change of interest rather than rigid motion, with no coherent motion field and no brightness constancy to exploit.
Second, we couple the encoding with a multi-task objective that returns infected-versus-mock status and a continuous time post-infection estimate from a single forward pass; the auxiliary objective lowers the time-estimation error, and we detected no accuracy difference on detection under this protocol.
Third, under leave-one-row-out cross-validation across three experimental runs, with the well as the unit of analysis, the encoding is more accurate than its single-frame counterpart by $+4.2$ percentage points over the whole time course and by $+23.9$ points in the late phase (40--48\,h), where the difference between the two procedures was largest. In matched fixed-split comparisons, the temporal advantage was observed with the three convolutional backbones, whereas the ViT-B/16 comparison showed no corresponding gain.
Fourth, we chart when reliable detection becomes available as a function of the multiplicity of infection (MOI) and time post-infection.
All results were obtained in a single cell preparation, a single virus strain, one imaging platform, and one laboratory, and are stated for that scope.
Because labels are assigned at the well level based on the experimental infection condition, the model should be interpreted as detecting image fields originating from infected versus mock wells, rather than directly measuring the infection status of individual cells.

\section{Related work}
\label{sec:related}

Two lines of work bear directly on this study: the encoding of temporal information into the color channels of a 2D network, which is where the representation used here comes from, and label-free imaging of infection and cell dynamics, which is the application. Table~\ref{tab:related} summarizes the closest methodological and application-specific precedents, and Section~\ref{sec:whatsnew} defines the methodological contribution and scope of the present study.

\subsection{Encoding temporal information into image channels}

Compact representations of temporal information in a single multi-channel image have been explored in several computer-vision settings; the first group of Table~\ref{tab:related} summarizes the closest methodological precedents.

Recent video models use dedicated architectures or pretraining objectives to learn spatio-temporal representations: video transformers attend over space and time jointly or in factorized form \cite{ref-bertasius2021timesformer,ref-arnab2021vivit}, and masked-autoencoder pretraining does so by masking a large fraction of the video patches and reconstructing them \cite{ref-tong2022videomae}. Self-supervised \cite{ref-jiang2023ssvqa} and masked-autoencoder \cite{ref-hu2026dsmae} pretraining have also been developed for video quality assessment, and for omnidirectional video, frame features adjusted by causal intervention for a projection-related confounder have been combined with temporal modeling across frames \cite{ref-hu2024ciqnet}. A simpler alternative rearranges several frames into one image for a 2D network, either as a grid of frames on which an image classifier is trained directly \cite{ref-fan2022sifar} or as the color channels of a single image \cite{ref-kim2022grayst}.

The representation used in the present study follows the temporal channel-stacking strategy described by Kim et al.: three time-offset grayscale frames stacked as R, G and B and passed to an ImageNet-pretrained 2D CNN. The present study differs primarily in the temporal and biological regime in which that representation is evaluated.

\subsection{Label-free detection of infection and cellular dynamics}

Deep learning on label-free microscopy is now established for cell segmentation, phenotyping and the prediction of fluorescent labels from transmitted light \cite{ref-christiansen2018insilico,ref-ounkomol2018labelfree,ref-moen2019dlmicroscopy}, and has been applied to cytopathic effect (CPE) and to viral infectivity by several groups. Andriasyan et al.\ predicted virus infection and revealed mechanisms of lytic infection from bright-field images \cite{ref-andriasyan2021iscience}; Petkidis et al.\ built an automated pipeline quantifying virus infectivity by label-free light microscopy \cite{ref-petkidis2024natcomm}; Chen et al.\ detected the CPE induced by influenza, parainfluenza and enterovirus with a convolutional network \cite{ref-chen2022biomedicines}; and Akkutay-Yoldar et al.\ described a web-based system for label-free virus classification and CPE detection \cite{ref-airvic2025}. These systems operate on individual frames. Chen et al.\ are also the closest precedent for the training objective used here: they trained a ResNet-50 in both single-task and multi-task form and found the multi-task model the more accurate of the two, with the advantage largest when training images for one virus were scarce \cite{ref-chen2022biomedicines}. Accordingly, the methodological distinction here lies not in multi-task learning itself but in coupling infection classification with continuous time regression. Their auxiliary tasks are all categorical --- virus identity across several host cell lines, routed by multiplexer and de-multiplexer layers --- whereas the second task here is a continuous regression onto time post-infection. This distinction is functionally important: a second categorical label partitions the same discrete decision, while a continuous target supervises the rate and stage of change, and it is what allows one forward pass to return both a call and an estimate of how far infection has progressed.

Related work on time-resolved biological imaging is summarized in the second group of Table~\ref{tab:related}. None of those studies uses the channel-stacking representation examined here.

In holographic flow cytometry, another label-free setting, Ciaparrone et al.\ noted that biases arising from different imaging settings and processing approaches may limit the generalization of learned models, and proposed to overcome them by extracting features with a convolutional autoencoder trained on unlabeled data before single-cell classification \cite{ref-ciaparrone2024unbiased}.

\subsection{Methodological contribution and study scope}
\label{sec:whatsnew}

The contribution can be separated into the representation, the experimental regime, and the empirical finding.

\emph{Representation.} Temporal channel stacking for an unmodified 2D CNN has previously been described and validated for action recognition \cite{ref-kim2022grayst,ref-bilen2018dynamicimages} and is used in multi-temporal remote sensing \cite{ref-daudt2018siamese}.

\emph{Empirical regime.} Those methods operate on human motion at sub-second to second offsets, where the between-channel difference is displacement of a moving subject. Here the offsets are three and six hours and the change of interest is biological progression --- cell rounding, shrinkage and detachment in a stationary monolayer, with no coherent motion field and no brightness constancy. Whether a representation designed for the first regime carries over to the second is not settled by the prior work, and testing it is one contribution of this study.

\emph{Biological scope.} We evaluate this question in primary human brain microvascular endothelial cells, a critical cellular component of the blood--brain barrier, infected with VEEV TC-83, the attenuated vaccine strain of an alphavirus whose virulent forms are Category B select agents and are neuroinvasive \cite{ref-weaver2004veev,ref-schafer2011vee}. This differs from the influenza, parainfluenza, enterovirus, adenovirus and rhinovirus systems in which label-free CPE detection has been developed \cite{ref-chen2022biomedicines,ref-petkidis2024natcomm,ref-andriasyan2021iscience,ref-airvic2025}, and the cytopathic kinetics of an alphavirus in endothelium need not match theirs. The present study therefore evaluates the encoding in a biological setting where it had not been examined.

\emph{Observed finding.} The benefit of temporal encoding is temporally localized rather than uniform: single-frame analysis degrades specifically in the late phase, and the temporal encoding does not. To our knowledge this late-phase divergence has not previously been reported for label-free CPE detection, and it is the difference that makes a trajectory representation preferable to a static one in this setting rather than merely equivalent to it.

\begin{table*}[t]
  \centering
  \caption{Prior work bearing on this study, in two groups: compact multi-channel encodings of time, which is where the representation comes from, and time-resolved label-free imaging of cells and pathogens, which is the application. The representation is shared with Kim et al.; the domain, the time scale and the reported phenomenon are not. The second group is listed because it is the nearest prior work on the application, not because it uses this encoding: none of those three does.}
  \label{tab:related}
  \footnotesize
  \setlength{\tabcolsep}{5pt}
  \renewcommand{\arraystretch}{1.15}
  \begin{tabular}{>{\raggedright\arraybackslash}p{0.155\linewidth}%
                  >{\raggedright\arraybackslash}p{0.215\linewidth}%
                  >{\raggedright\arraybackslash}p{0.155\linewidth}%
                  >{\raggedright\arraybackslash}p{0.085\linewidth}%
                  >{\raggedright\arraybackslash}p{0.245\linewidth}}
    \toprule
    \textbf{Work} & \textbf{Temporal encoding} & \textbf{Domain} & \textbf{Offsets} & \textbf{Relation to the present study} \\
    \midrule
    \multicolumn{5}{l}{\emph{Compact multi-channel encodings of time}} \\
    Bilen et al.~\cite{ref-bilen2018dynamicimages} & Rank pooling of a video into one RGB image & Human action recognition & $\sim$1\,s & Derived summary, not raw frames; different domain and time scale \\
    Kim et al.~\cite{ref-kim2022grayst} & Three time-sampled grayscale frames as R, G, B; one ResNet50 backbone, five temporal frameworks & Human action recognition & $\sim$0.1--1\,s & \textbf{Same representation}; differs in domain, time scale, and the phenomenon reported \\
    Caye Daudt et al.~\cite{ref-daudt2018siamese} & Multi-date composites across channels & Remote-sensing change detection & Days--years & Different domain; change is scene-level, not morphological \\
    \midrule
    \multicolumn{5}{l}{\emph{Time-resolved label-free imaging of cells and pathogens}} \\
    Paquin et al.~\cite{ref-paquin2022colonies} & Spatio-temporal features over incubation & Bacterial colony ID & Hours & No channel stacking; bacterial growth, not viral CPE \\
    DynaCLR \cite{ref-dynaclr2024} & Contrastive learning with temporal regularization & Cell and infection dynamics & Hours & Representation learned by a dedicated objective, not a fixed encoding \\
    Liu et al.~\cite{ref-liu2023plaque} & Spatio-temporal signal from holograms & Viral plaque quantification & Hours & Different imaging modality and endpoint \\
    \bottomrule
  \end{tabular}
\end{table*}

\section{Materials and methods}

\subsection{Materials and reagents}
\label{sec:keyres}
Primary human brain microvascular endothelial cells (HBMVEC) were obtained from Cell
Systems (an AnaBios company; Cat\#ACBRI~376, Lot\#376.07.05.01.2F) and cultured in EGM-2 Endothelial Cell Growth Medium-2 BulletKit (Lonza, Cat\#CC-3162) supplemented with 5\% fetal bovine serum and 1\% penicillin/streptomycin. Vero cells (African green monkey kidney epithelial, ATCC Cat\#CCL-81), used for the plaque assay, were cultured in Dulbecco's Modified Eagle Medium supplemented with 10\% fetal bovine serum and 1\% penicillin/streptomycin. All cells were maintained at 37$^{\circ}$C and 5\% CO$_2$ in a humidified incubator. The attenuated Venezuelan equine encephalitis virus vaccine strain TC-83 was a
P1 laboratory stock generated from an infectious cDNA clone originally provided by
I.~Frolov. Reagents were Dulbecco's phosphate-buffered saline (Corning,
Cat\#21-030-CV), fetal bovine serum (Gibco, Cat\#A5256801), Dulbecco's Modified Eagle
Medium (Quality Biological, Cat\#112-014-101CS), 2$\times$ Minimum Essential Medium
without phenol red (Quality Biological, Cat\#115-073-101), trypsin
(0.25\%)--EDTA (0.02\%) (Quality Biological, Cat\#118-093-721), penicillin/streptomycin (Corning, Cat\#30-002-CI), agarose, 10\% formalin (Cat\#H501128-4L), crystal violet (Cat\#405835000) and ethanol (Cat\#BP2818-4). Cells were seeded in 12-well plates
(CELLSTAR, Greiner Bio-One, Cat\#665180). Imaging used the CELLCYTE~X live-cell system
with CELLCYTE Studio (CYTENA GmbH, Freiburg, Germany). Analysis used Python~3.10,
PyTorch~2.0 and torchvision, scikit-learn, OpenCV, and
GraphPad Prism v11.0.0. The analysis pipeline is available at
\url{https://github.com/CAIR-LAB-WFUSM/cell_tempo}.

\subsection{Experimental model and study participant details}

\subsubsection{Cell lines}
\label{sec:celllines}

The commercially obtained HBMVEC preparation named in Section~\ref{sec:keyres} was cultured at 37$^{\circ}$C and 5\% CO$_2$ in the medium named there.
Per the vendor certificate of analysis, the preparation tested negative for HIV, HBV and HCV; was authenticated by short tandem repeat (STR) profiling; passed independent sterility testing for bacterial, fungal and mycoplasma contamination; and was validated as endothelial by immunofluorescence for CD31, von~Willebrand Factor / Factor~VIII, and cytoplasmic uptake of Di-I-Ac-LDL, each $>$95\% positive.
Post-thaw viability and recovery were high, with viability at receipt reported as 92\%; the donor was male and pediatric.
Working cell banks were prepared from the parental vial according to the manufacturer's instructions.
Cells were used only within the manufacturer's recommended passage range and at $>$85\% viability at each passage; the three experimental runs analyzed here were performed at passages~4, 5 and 6, one passage per run.
Viability was confirmed at every passage, but the individual per-passage values were not retained, so per-passage viability cannot be reported here.
No sex-based comparisons were performed in this study, which used a single commercially obtained primary HBMVEC preparation as an in vitro model of the brain endothelium.

\subsubsection{Virus strains}

The attenuated Venezuelan equine encephalitis virus (VEEV) vaccine strain TC-83 was used for all infection experiments.
Virus inoculum was prepared from P1 laboratory stock generated from an infectious cDNA clone originally provided by I.~Frolov \cite{ref-kinney1993tc83} and maintained under approved institutional biosafety procedures.
All work involving VEEV TC-83 was conducted under BSL-2 containment under institutional biosafety committee protocol SPROTO202300000019~(22-06).
No human subjects or animals were involved; the HBMVEC preparation was obtained commercially.

\subsection{Method details}

\subsubsection{Viral infection protocol and endpoint titration}

HBMVEC were seeded at a density of 45,000 cells per well in 12-well tissue-culture plates to obtain a confluent monolayer suitable for morphological assessment.
VEEV TC-83 was used for infection at three multiplicities of infection: MOI~5, MOI~1, and MOI~0.1. These MOI levels were selected to span a range of CPE kinetics from rapid synchronous infection (MOI~5) to delayed heterogeneous responses (MOI~0.1).
A mock-infected control condition was treated with PBS in parallel.
After viral adsorption, the HBMVEC monolayers were washed with phosphate-buffered saline (PBS), and fresh medium was added (defined as the start of the experiment, $t=0$).

Terminal supernatants were collected at the end of the 48-hour imaging experiment and stored at $-80^{\circ}$C until titration. Infectious titers were determined by plaque assay on Vero monolayers. Vero cells were seeded at 180,000 cells per well in 12-well plates and grown overnight to approximately 90\% confluency. Supernatants were serially diluted ten-fold in DMEM with 1\% penicillin/streptomycin and added to the monolayers in triplicate, at an inoculum volume of 100\,$\mu$L per well. Plates were incubated for 1 hour at 37$^{\circ}$C and 5\% CO$_2$, with rocking approximately every 5 minutes, then overlaid with 1\,mL of 0.6\% agarose diluted 1:2 in 2$\times$ EMEM supplemented with 5\% FBS and 1\% penicillin/streptomycin, for a final agarose concentration of 0.3\%. After a further 48 hours at 37$^{\circ}$C and 5\% CO$_2$, plates were fixed with 10\% formalin for at least 1 hour, the agarose plugs were removed, and monolayers were stained with 1\% crystal violet in 20\% ethanol for 15 minutes, then washed with running water, dried, and the plaques counted. Titers are reported as PFU/mL, computed as the mean plaque count per well multiplied by the dilution factor and by ten, which converts the 100\,$\mu$L inoculum to a per-milliliter basis. The limit of detection is one plaque per 100\,$\mu$L inoculum, or 10\,PFU/mL. Three technical replicates were run per condition per experimental run; results are shown in Figure~S7.

\subsubsection{Experimental design and unit of analysis}

Three independent experimental runs were performed on separate days and are referred to here as Run~1, Run~2, and Run~3 for presentation clarity. They are independent preparations, infections and acquisitions, but all three derive from the same commercially obtained HBMVEC lot rather than from three donors and therefore do not represent three independent biological sources.
Each experimental run used a separate 12-well plate with four conditions (MOI~5, MOI~1, MOI~0.1, and Mock) and three replicate wells per condition (rows A, B, and C).
Each well was imaged at 36 acquisition positions in a 6$\times$6 grid pattern over 96 time points collected every 30 minutes across 48 hours.
The acquisition hierarchy is therefore run $\rightarrow$ well $\rightarrow$ position $\rightarrow$ timepoint $\rightarrow$ temporal pseudo-RGB sample (Figure~S6): 3 runs, 12 wells per run, 36 positions per well, and 96 frames per position.

The well is the unit of analysis in this study, and the study contains 36 wells.
The 36 positions within a well image the same physical culture under the same treatment and are not spatially independent; the 96 frames acquired at one position are a time series of a single field and are strongly autocorrelated.
The wells are themselves nested within the three runs, 12 to a run, so they are not 36 mutually independent biological units either; the inference below is conditional on these three runs, and Section~\ref{sec:stats} reports the effect separately within each of them.
The frame-level count implied by the acquisition schedule, $3 \times 12 \times 36 \times 96 = 124{,}416$, is therefore reported as a measure of acquisition volume and not as a number of independent observations.
All partitioning, resampling, and significance testing reported in this study is performed with the well as the unit, and every accuracy quoted for a well is computed over that well's frames before the well enters any test.

\subsubsection{High-frequency live-cell imaging}

Imaging was performed using the CELLCYTE X (CYTENA GmbH, Freiburg, Germany) live-cell imaging system located inside a standard incubator.
Images were acquired using a 10$\times$ objective in ``Enhanced Contour'' mode at a spatial resolution of 0.345~$\mu$m/pixel.
Each field of view was 2448$\times$2048 pixels ($\sim$0.8$\times$0.7~mm), and 36 such fields were acquired per well in a 6$\times$6 grid.
Images were acquired every 30 minutes over 48 hours, producing per-position time series stored as multi-frame TIFF stacks.
The 10$\times$ objective was chosen for contrast rather than for coverage: it resolves the cell-shape changes that constitute CPE, at the cost of sampling only part of the well area, which contributes to the between-well spread discussed in Section~\ref{sec:limitations}.

\subsubsection{Cell-culture and image quality control}
\label{sec:qc}

Quality control comprised four complementary procedures.

First, the vendor certificate of analysis for the cell preparation, summarized in Section~\ref{sec:celllines}.
Second, routine culture practice: working cell banks prepared from the parental vial, use restricted to the manufacturer's recommended passage range, and $>$85\% viability confirmed at each passage, with the limitation on per-passage records noted above.
Third, image-based confluency evaluation, performed at the imaging site on every well of every experimental run and inspected for inter-well variability before analysis.
This measure is intrinsically variable, because the segmentation mask was required to capture cells both as an adherent healthy monolayer and as they rounded up and detached following infection; the same mask must therefore accept two very different appearances, and the resulting confluence values spread accordingly.
Fourth, image-level review before analysis: acquired fields were inspected visually for focus, debris, bubbles, and uneven illumination, and cell numbers were estimated independently with Cellpose~\cite{ref-stringer2021cellpose} as a check that fields contained the expected culture density.
No image, position or well was excluded on the basis of this review, and no automated image-quality filter was applied: every acquired frame from every well enters the analysis.

\subsubsection{Temporal pseudo-RGB encoding}

To incorporate temporal context, each sample is represented as a three-channel image composed of frames at three time offsets relative to the current time point $t$:
\begin{equation}\label{eq:pseudorgb}
  I_{\mathrm{pseudo}}(t) = \bigl[\,f(t{-}6\text{h}),\;\; f(t{-}3\text{h}),\;\; f(t)\,\bigr],
\end{equation}
where $f(\tau)$ denotes the normalized grayscale frame at time $\tau$.
Each raw grayscale frame is first rescaled to [0,~1] using per-frame min--max normalization by subtracting the frame minimum and dividing by the resulting dynamic range (equivalently, the post-subtraction maximum), and then quantized to 8-bit; after stacking the three time-offset frames as a pseudo-RGB image, channel-wise standardization is applied with mean $=[0.5, 0.5, 0.5]$ and standard deviation $=[0.25, 0.25, 0.25]$ before the tensor is passed to the backbone.
The three channels share these constants, rather than the per-channel ImageNet statistics, because the channels hold one grayscale modality at three time points rather than three colors: shared constants center the [0,~1] range of each frame on zero and standardize all three frames identically, so that a difference between channels reflects a difference between the normalized frames.
The same per-frame normalization and the same standardization constants are applied identically in both the temporal and single-frame configurations, so any performance difference reflects temporal encoding rather than preprocessing.
The red channel maps to the oldest frame ($t{-}6$\,h), the green channel to an intermediate frame ($t{-}3$\,h), and the blue channel to the current frame ($t$).
Regions undergoing morphological change appear as color shifts, while static regions appear gray; Figure~S2 shows representative encoded samples.
This allows a standard ImageNet-pretrained CNN to process temporal differences encoded across channels without architectural modification.

For frames near the experiment start where offsets reach before $t=0$, the earliest available frame is used (clamp padding): $f(\tau) = f(\max(0, \tau))$.
As a baseline, the single-frame model replicates the same normalized grayscale frame across all three channels, so the two models differ only in the two earlier frames supplied to the temporal model.

Per-frame normalization has a specific sensitivity that follows from the construction.
Because each channel is scaled by the extrema of its own frame, a transient artifact confined to one offset --- debris entering the field, a bubble, a focus excursion --- rescales that channel alone and therefore appears as a global color cast across the whole field rather than as a localized change.
This failure mode is specific to the temporal configuration: in the single-frame configuration all three channels are copies of one frame, so any such artifact rescales all three identically and no color cast can arise.
Section~\ref{sec:normsens} reports a retraining experiment under an alternative percentile-based normalization that bounds the influence of single-frame extrema.

The three offsets ($-$6, $-$3, 0\,h) were fixed before any model was trained and were not tuned.
The comparison across alternative offsets reported in Section~\ref{sec:ablation} is a post hoc sensitivity analysis on the held-out partition; no configuration reported elsewhere in this paper was selected on the basis of it.

\subsubsection{Dataset construction and preprocessing}

Data were organized into four conditions based on MOI level.
Each frame was assigned a four-class label based on experimental condition ($y_{\mathrm{cls}} \in \{0, 1, 2, 3\}$ for MOI~5, MOI~1, MOI~0.1, and Mock).
Binary labels were derived post-hoc by collapsing classes 0--2 into ``Infected.''
A deterministic 5\%-per-side center crop was applied to mitigate potential overlap between adjacent acquisition positions ($\sim$5\%), followed by resizing to 512$\times$512 pixels.

Per-frame normalization, crop, and resize are deterministic functions of the raw frame, so this prefix of the pipeline was precomputed once into an 8-bit cache to make the full experimental grid tractable.
The cache was verified against the on-the-fly path over samples drawn from six stacks at three frame indices, including clamp-padded and terminal frames: maximum absolute difference 0 and correlation 1.000000, i.e.\ bit-exact.
Augmentation remains at training time and is never cached.

Augmentation was applied to training data only and comprised random horizontal and vertical flips together with rotation by a random multiple of 90$^{\circ}$.
Rotation by a multiple of 90$^{\circ}$ on a square field is a lossless permutation of pixels: it requires no interpolation and introduces no boundary padding, and microscopy fields have no canonical orientation.
It was adopted on that ground, and the same augmentation is applied identically in every configuration, backbone and baseline reported here.

\subsubsection{Evaluation protocol}
\label{sec:protocol}

Three partitioning schemes are used, and each result in this paper states which applies.
All three separate training from test data at the level of the physical well, so that no model is evaluated on a well it was trained on.

\emph{Fixed row split.} Rows A and B --- two of the three replicate wells per condition within each experimental run --- were used for training, and row C was held out for testing.
The held-out partition consists of 12 wells across the three experimental runs and 41,472 frame-level samples.
This is the protocol for the ablation and the temporal-offset sensitivity analysis (Section~\ref{sec:ablation}), the baseline comparison (Section~\ref{sec:baselines}), and the backbone comparison (Section~\ref{sec:backbones}), in all of which the quantity of interest is a difference between arms evaluated on identical data.

\emph{Leave-one-row-out cross-validation.} Because the study contains only three plates and one row of each is held out, a single split leaves the result dependent on which row that is.
The protocol was therefore repeated as three-fold cross-validation in which each fold holds out one row across all three plates.
Every well is held out exactly once, the train/test proportion is constant across folds, and pooling the out-of-fold predictions yields one prediction per well from a model that never saw that well in training, for all 36 wells.
This is the protocol for the paired significance tests and balanced binary metrics (Section~\ref{sec:stats}), as well as the per-fold results (Section~\ref{sec:cv}).

\emph{Leave-one-run-out cross-validation.} As an additional within-study robustness check across experimental runs, the same protocol was repeated with the folds defined by run rather than by row, so that each fold holds out all twelve wells of one run and the model is evaluated on a plate not represented in training. This scheme and the row-out scheme both train on 24 wells and test on 12, so they differ in what is withheld rather than in how much. It is reported in Section~\ref{sec:cv} and is secondary to the leave-one-row-out analysis, which remains the protocol for the headline result.

In all three schemes wells are assigned by a grouping variable fixed in advance --- plate row or experimental run --- rather than individually, so no reported partition depends on a per-well choice.

\subsubsection{Multi-task architecture and training}
\label{sec:training}

ResNet50 was selected as the primary backbone because Chen et al.\ used the same backbone for single-task and multi-task CPE detection \cite{ref-chen2022biomedicines}, providing an architecture-level connection to prior CPE work rather than a direct cross-study performance comparison.
The shared feature encoder is a torchvision ResNet50 \cite{ref-he2016resnet} initialized with ImageNet \cite{ref-deng2009imagenet} weights, with its final fully connected layer replaced by an identity mapping, so that the encoder emits the 2,048-dimensional globally average-pooled feature that follows the final residual stage.
Two task heads consume that feature, each with the structure
\begin{equation}\label{eq:heads}
  \begin{aligned}
    &\mathrm{Dropout}(0.2) \rightarrow \mathrm{Linear}(2048 \rightarrow 256) \rightarrow \mathrm{ReLU} \\
    &\qquad \rightarrow \mathrm{Dropout}(0.2) \rightarrow \mathrm{Linear}(256 \rightarrow k),
  \end{aligned}
\end{equation}
with $k=4$ for the condition-classification head (MOI~5, MOI~1, MOI~0.1, Mock) and $k=1$ for the time-regression head.
The primary binary readout is derived from the four-class head by collapsing classes 0--2, and is not a separate output.
All backbone layers are fine-tuned: no layers are frozen, and no discriminative or layer-wise-decayed learning rates are used.
An optional mechanism in the implementation that conditions the regression head on the classification logits was disabled in every run reported here.

The loss function is:
\begin{equation}\label{eq:loss}
  L = \lambda_{\mathrm{cls}} \cdot L_{\mathrm{CE}} + \lambda_{\mathrm{reg}} \cdot L_{\mathrm{SL1}},
\end{equation}
where $L_{\mathrm{CE}}$ is cross-entropy loss and $L_{\mathrm{SL1}}$ is Smooth L1 (Huber) loss, with $\lambda_{\mathrm{cls}} = \lambda_{\mathrm{reg}} = 1.0$.
For single-task ablation experiments, the weight of the excluded task was set to zero. A classification-only model therefore has no regression output, and a regression-only model has no trained classification head at all: the head is instantiated but receives no gradient, so its outputs are not a readout of anything and the corresponding metrics are reported as not applicable rather than as a number. The regression-only runs were launched from the binary configuration, which fixes only which head is instantiated and has no bearing on the model that results.

Every model we trained ourselves uses one frozen training protocol, identical across the arms, the backbones and the internal baselines, though not the published comparator, which keeps its own recipe (Section~\ref{sec:baselineimpl}): 30 epochs, AdamW \cite{ref-loshchilov2019adamw} with learning rate $1\times10^{-4}$ and weight decay $1\times10^{-4}$, cosine annealing \cite{ref-loshchilov2017sgdr} with $T_{\max}=30$ and $\eta_{\min}=10^{-6}$, batch size 96, automatic mixed precision, gradient-norm clipping at 1.0, and 512$\times$512 input.
These values were carried over from prior experience with the same backbone and imaging modality and were not tuned for this study.
No hyperparameter search was performed, over these or any other training parameter; consequently no held-out data was consumed by model selection, and the epoch budget was not tuned.

The held-out partition was evaluated once per run, at the terminal epoch.
Because the number of epochs and the annealing schedule were fixed in advance and the schedule completes at epoch 30, the reported checkpoint is the endpoint of a predetermined schedule rather than a checkpoint chosen by observing test performance.
No early stopping was used and no model, epoch, or configuration was selected on held-out performance.
Learning curves built from the training history alone are reported in the supplement (Figure~S8); no held-out quantity appears in them.

Each configuration we trained was run with five random seeds (42--46), and its values are reported as the mean $\pm$ the standard deviation across those seeds. The non-deep baselines are deterministic given the split and were run once; the pooled out-of-fold and paired quantities are defined in Section~\ref{sec:uncert} and are not seed means.

\subsubsection{Backbone comparison}

To test whether the encoding depends on the specific backbone, the identical protocol was run with and without the temporal encoding on four ImageNet-pretrained architectures: ResNet50 \cite{ref-he2016resnet}, EfficientNet-B0 \cite{ref-tan2019efficientnet}, ConvNeXt-Tiny \cite{ref-liu2022convnext}, and ViT-B/16 \cite{ref-dosovitskiy2021vit}, all taken from torchvision with default pretrained weights and differing only in the encoder.
ViT-B/16 was included because it differs from the other three in how the input channels are first combined: a convolutional network applies many small filters across the channels at every spatial position and at several scales, whereas a vision transformer combines them once, inside a single linear projection of each $16\times16$ patch. Video architectures address that difference with purpose-built temporal operators, whether a shift along the channel axis of a 2D network \cite{ref-lin2019tsm} or attention factorized over space and time \cite{ref-bertasius2021timesformer}, neither of which is available to a plain ImageNet-pretrained ViT given three stacked frames. Whether the difference matters for this encoding is an empirical question, and it is the reason a transformer was included rather than a fourth convolutional network.

\subsubsection{Baseline implementations}
\label{sec:baselineimpl}

The learned baselines other than the published comparator were trained under the frozen protocol of Section~\ref{sec:training}, on the same split, with the same heads and losses, and differ only in how temporal information reaches the network. The comparator is the exception and is described separately below: reproducing a published method means adopting parts of its training recipe and not only its architecture.

\emph{Optical flow.} Dense optical flow was computed with the Farneb\"{a}ck algorithm \cite{ref-farneback2003flow} as implemented in OpenCV. Flow was computed between \emph{consecutive} 30-minute frames and accumulated across the window, rather than directly across the 3--6\,h offsets, because the brightness-constancy and small-displacement assumptions on which flow estimation rests do not hold over hours of morphological change. This baseline therefore observes about 13 consecutive frames per sample and receives more input than the proposed encoding, not the same input; this asymmetry favors the baseline.

\emph{Recurrent model.} A ResNet50 encoder applied frame-wise to the same three time-offset frames, followed by an LSTM over the resulting sequence of features and the same two heads \cite{ref-donahue2015lrcn}.

\emph{3D convolutional network.} A torchvision \texttt{r3d\_18} spatiotemporal residual network applied to the same three frames as a short volume \cite{ref-tran2015c3d}.

\emph{Published comparator.} The single-frame architecture of Petkidis et al. \cite{ref-petkidis2024natcomm} was reimplemented from their published description: an EfficientNet-B0 backbone trained from random initialization with their optimizer and augmentation. It therefore differs from our models in four respects at once --- architecture, initialization, optimizer and augmentation. Accordingly, this comparison evaluates the published procedure as a whole rather than isolating the effect of architecture. What is held constant is the data, the split, the evaluation and the 30-epoch budget: their procedure selects the epoch with the lowest validation loss and holds out whole plates, whereas we keep our own split and evaluate once at the terminal epoch, so that no part of the comparison depends on their model selection touching our held-out data. Because the network is randomly initialized rather than fine-tuned from pretrained weights, its run-to-run variance is larger than that of the pretrained arms.

\emph{Non-deep baselines.} Three representations were reduced to scalar summary statistics per sample and classified without any convolution or pretrained weights: absolute frame differences at the 3\,h and 6\,h offsets, the accumulated Farneb\"{a}ck flow magnitude, and a confluence estimate with its 3\,h and 6\,h changes and its local slope. For the flow features here the magnitudes were recomputed in floating point rather than read from the quantized training cache. Each representation was classified at two levels: a single cut-off, with both the threshold and its direction fitted on the training partition rather than assumed, and a logistic regression on the full feature vector. The cut-off form involves no fitted weights beyond the threshold itself; the logistic regression is a shallow learned classifier and is described as such.

\subsubsection{Positional control}
\label{sec:positional}

The condition-to-column mapping was identical on all three plates, so at the level of the well, column is condition and row is the cross-validation fold: no well-level stratification can separate positional effects from condition effects. We therefore evaluate positional variation within each well. The 36 acquisition positions form a 6$\times$6 grid, of which the outer ring of 20 lies nearer the well wall and the inner 16 do not, and within a well the condition, the run and the fold are all held constant by construction, so position is the only thing that varies. For each held-out well we therefore computed accuracy over its edge positions and over its interior positions separately and tested the paired within-well difference over the 36 wells with the same cluster bootstrap used elsewhere. The same comparison restricted to the nine mock wells is also reported, since those contain no infection signal and any structure across position in them must be positional. The outer ring is the same set of 20 positions under row-major, column-major, row-serpentine and column-serpentine numbering, so the analysis does not depend on which of those the acquisition software used.

\subsubsection{Normalization sensitivity analysis}
\label{sec:normsens}

To test whether the results depend on the per-frame min--max normalization, both the temporal and the single-frame models were \emph{retrained} from scratch under an alternative normalization in which each frame is scaled by its 1st and 99th percentiles rather than its extrema, bounding the influence of single-pixel outliers.
Retraining, rather than re-evaluating the existing models on renormalized images, is necessary because the latter measures robustness to an inference-time distribution shift and not the consequence of the normalization choice itself.
The comparison is reported as the paired difference between the temporal and single-frame arms under each pipeline.

\subsection{Quantification and statistical analysis}

\subsubsection{Endpoints and metrics}

To avoid ambiguity between the four-class training target and the binary endpoint, we report two related classification outputs: the primary binary infected-versus-mock readout and the auxiliary four-class condition readout.
The condition-classification head predicts four experimental conditions (MOI~5, MOI~1, MOI~0.1, and Mock), and the primary binary infected-versus-mock readout is derived by collapsing the condition-classification predictions.

For the binary endpoint, the positive class is Infected.
Sensitivity, specificity, precision, F1, AUROC, and AUPRC \cite{ref-fawcett2006roc,ref-davis2006pr} are reported with that convention throughout.
Because three of the four conditions are infected, the binary task carries a 3:1 class ratio and a majority-class accuracy of 75\%; that value, not 0.5, is the reference any binary accuracy in this paper must be read against, and it is why balanced metrics are reported alongside accuracy (Section~\ref{sec:balanced}).
The four-class task is balanced by design, with equal numbers of wells per condition.
The regression target is the elapsed time since inoculation, defined identically for all four conditions: for the three infected conditions it is time post-infection, and for Mock it is time since the parallel mock treatment, so the target exists for every sample, and Mock is not assigned an infection age it does not have. It was evaluated using mean absolute error (MAE), root mean square error (RMSE), and coefficient of determination ($R^2$).
All metrics were additionally computed in 3-hour time bins across the 0--48\,h timeline for temporal analysis.

\subsubsection{Uncertainty and significance}
\label{sec:uncert}

Two sources of variability are quantified separately, because they answer different questions.
Variability across training runs is quantified by repeating every reported configuration with five random seeds and reporting the mean and the standard deviation across seeds.
Between-well variability within the observed runs is quantified by resampling wells. Because the 36 wells come from one cell lot and are nested in three runs, this resampling does not quantify variability across independent biological material.

For the fixed-split results, bootstrap 95\% confidence intervals were computed by resampling the 12 held-out wells with replacement 1,000 times; within each resample all frame-level predictions from the selected wells were aggregated and the metric recomputed, and the 2.5th and 97.5th percentiles of the resulting distribution define the interval.

For the paired comparison between the temporal and single-frame models, the unit is the well and all 36 out-of-fold wells contribute exactly once.
Each well's accuracy was averaged over the five seeds before entering the test, so that the comparison is between two procedures rather than between two particular fitted models.
Three complementary tests are reported on the same 36 paired differences: a cluster bootstrap over wells \cite{ref-field2007cluster} with 10,000 resamples, giving the interval; a sign-flip permutation test evaluated by Monte Carlo over $10^5$ sign assignments, so the smallest attainable p-value is $1.0 \times 10^{-5}$; and a Wilcoxon signed-rank test \cite{ref-wilcoxon1945}, which uses the signs and ranks of the absolute nonzero paired differences and, read as a test of location, assumes that the distribution of paired differences is symmetric about zero under the null.

All three tests treat the well as the exchangeable unit. Because wells are nested within three runs, this is conditional inference given those runs and does not propagate run-to-run variability; with three runs a random-effect term for run cannot be estimated with any precision, so we report the effect within each run instead and treat agreement across the three as the evidence that the result is not carried by one plate.

The baseline comparisons of Section~\ref{sec:baselines} are on the fixed row split, so their paired tests resample the 12 held-out wells rather than the 36 out-of-fold wells. With 12 wells the sign-flip reference distribution has only $2^{12} = 4096$ members, so it is enumerated exhaustively rather than sampled and the resulting p-values are exact, with $2/4096 = 0.0005$ the smallest attainable. Five comparisons against the proposed encoding are made at once and are reported with a Holm correction \cite{ref-holm1979}; their power at 12 wells is low by construction. The comparison between joint and regression-only training on the time-estimation endpoint is a separate test on a different quantity, is reported in hours, and is not part of that family of five, so no correction is applied to it.

The late-phase endpoint is the final eight hours of the time course, 40--48\,h. This interval was fixed before any of the analyses reported here, as the window in which single-frame appearance becomes ambiguous, and so was chosen independently of the results below; the study was not formally preregistered. The whole-timeline comparison is reported alongside it, under the same leave-one-row-out protocol, over the same 36 out-of-fold wells and with the same paired tests (Table~\ref{tab:4}).

Frame-level significance testing --- for instance McNemar's test over the roughly $1.2\times10^{5}$ frames --- is deliberately not used for inference. Frames within a position are strongly autocorrelated and positions within a well are not independent, so such a test treats correlated observations as independent, understates the standard error, and would return an implausibly small p-value for an effect estimated from 36 wells. Where frame counts appear in this paper they describe acquisition volume only.

Most quantitative result values in this paper appear in the tables, each of which states the protocol and the form of its values in its caption. The text also reports simple differences, ranges, counts and other summaries derived from tabulated values, as well as results from analyses for which no table is provided: the well-level paired results for joint versus regression-only training and for the fixed-split baselines; the run-by-run differences of Section~\ref{sec:cv}; the late-phase leave-one-run-out comparison and seed dispersion; and the within-well positional controls. The figures showing per-bin curves, confidence trajectories, confusion matrices and feature embeddings (Figures~\ref{fig:4} and~\ref{fig:2} and Supplementary Figures~S1, S3, S4 and~S9) are drawn from a single training run of the reported configuration on the fixed row split and are illustrative; they are not the source of any value quoted in the text.

Software: Python 3.10, PyTorch 2.0 \cite{ref-paszke2019pytorch}, torchvision, scikit-learn \cite{ref-pedregosa2011sklearn}, OpenCV.
t-SNE visualizations \cite{ref-vandermaaten2008tsne} used scikit-learn with perplexity~30 and 1,000 iterations on 2,048-dimensional backbone features.

\section{Results}

Unless stated otherwise, every value is the mean $\pm$ the standard deviation over five independent random initializations of the same frozen protocol, evaluated on the held-out row of each plate. Accuracies are given as percentages throughout, and standard deviations of an accuracy are therefore in percentage points; AUROC and AUPRC are reported on their natural 0--1 scale. Cross-validated and pooled results are reported in Section~\ref{sec:cv}.

\subsection{Contribution of temporal encoding and of multi-task supervision}
\label{sec:ablation}

Temporal encoding is the dominant contributor to infected-versus-mock detection (Table~\ref{tab:1}). With a binary head, replacing the single frame with the temporal pseudo-RGB input raises accuracy from $88.75 \pm 0.46$ to $93.26 \pm 0.44$ under classification-only training, and from $90.19 \pm 0.05$ to $93.24 \pm 0.28$ under multi-task training. The same holds with the four-class head, where the binary readout is derived by collapsing the three infected classes: $85.87 \pm 0.64$ to $92.60 \pm 0.35$ without the auxiliary task, and $88.38 \pm 2.40$ to $93.26 \pm 0.11$ with it. The encoding contributes between $3.0$ and $6.7$ percentage points in every one of these four comparisons.

Multi-task supervision behaves differently from the encoding. On the binary task, the present comparisons do not provide evidence that it improves classification: with the temporal input, classification-only ($93.26 \pm 0.44$) and multi-task ($93.24 \pm 0.28$) yielded similar mean accuracies across the five initializations. With the four-class head the multi-task arm is 0.66 percentage points ahead ($92.60 \pm 0.35$ versus $93.26 \pm 0.11$); that 0.66-point contrast was not tested at the well level and is therefore not interpreted as evidence of improved detection. On the regression the joint objective is ahead of a regression-only model trained on the same input --- MAE $= 1.20 \pm 0.04$\,h against $1.25 \pm 0.06$\,h, with a higher $R^2$ ($0.9852 \pm 0.0009$ versus $0.9784 \pm 0.0105$). Tested at the well level over the 12 held-out wells, the paired difference is $-0.053$\,h in favor of joint training (95\% CI $[-0.091, -0.020]$\,h; exact sign-flip $p = 0.013$), with 9 of the 12 wells favoring it. This effect is small and unevenly distributed across the three runs. It amounts to about 3 minutes against a time-estimation error of roughly 75 minutes, with an interval running from 1 to 5 minutes, and the per-run differences are --- $-0.131$, $-0.021$ and $-0.007$\,h --- so although the direction is the same in all three, excluding the largest of them leaves the difference unresolved ($p = 0.20$). The well-level analysis therefore resolves a small overall effect, but the present sample does not support resolving how the effect varies across runs. The clearest practical benefit of joint training is efficiency: one forward pass returns both a call and an estimate of how far the process has advanced, and neither readout requires a second model.

\begin{figure*}[t]
  \centering
  \includegraphics[width=\linewidth]{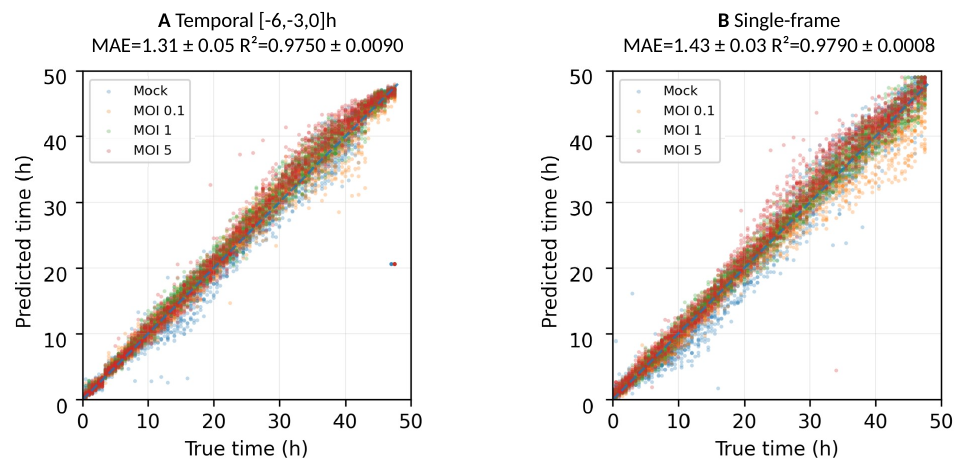}
  \caption{Time post-infection is recovered as a continuous secondary readout. Predicted vs.\ actual time post-infection for (A) the temporal multi-task model (MAE $= 1.31 \pm 0.05$\,h, $R^2 = 0.9750 \pm 0.0090$) and (B) the single-frame multi-task counterpart (MAE $= 1.43 \pm 0.03$\,h, $R^2 = 0.9790 \pm 0.0008$). Points are colored by infection condition (MOI~5, MOI~1, MOI~0.1, Mock), and each panel carries its own axes and legend; the dashed diagonal denotes perfect agreement. The points are from a single representative run on the fixed row split; the errors quoted above are the five-seed means of Table~\ref{tab:1}. The temporal model has the lower mean absolute error and the marginally lower $R^2$; neither temporal-versus-single-frame difference shown here was tested at the well level.}
  \label{fig:4}
\end{figure*}

Where the four-class readout errs is time-dependent: both models confuse the multiplicities early, and in the last window the single-frame model additionally calls Mock fields infected while the temporal model does not (Figure~S3). The configuration used throughout the remainder of the paper is the temporal four-class multi-task model: binary accuracy $93.26 \pm 0.11$, four-class accuracy $66.81 \pm 0.56$, MAE $= 1.31 \pm 0.05$\,h, $R^2 = 0.9750 \pm 0.0090$; its time-post-infection predictions are shown against the true elapsed time in Figure~\ref{fig:4}.
The temporal advantage is preserved across the evaluated spans: every offset configuration tested between 3 and 12\,h exceeds the single-frame baseline on all three metrics, and the reported 6\,h span has the highest mean on all three (Table~\ref{tab:2}). Differences among spans were not tested at the well level.

\begin{table*}[t]
  \centering
  \caption{Contribution of temporal encoding and of multi-task supervision. All models share one frozen protocol and the same fixed row split, and differ only in the input encoding, the output head, and the training objective. Accuracies are percentages; dashes denote outputs not defined for the corresponding model. Values are mean $\pm$ SD across five random initializations.}
  \label{tab:1}
  \footnotesize
  \setlength{\tabcolsep}{5pt}
  \renewcommand{\arraystretch}{1.15}
  \begin{tabular*}{\linewidth}{@{\extracolsep{\fill}}lllrrrr@{}}
    \toprule
    \textbf{Encoding} & \textbf{Classification head} & \textbf{Objective} & \textbf{Binary accuracy} & \textbf{4-class accuracy} & \textbf{MAE (h)} & \textbf{$R^2$} \\
    \midrule
    Single-frame & Binary & Classification only & $88.75 \pm 0.46$ & \multicolumn{1}{c}{--} & \multicolumn{1}{c}{--} & \multicolumn{1}{c}{--} \\
     & Binary & Multi-task & $90.19 \pm 0.05$ & \multicolumn{1}{c}{--} & $1.43 \pm 0.02$ & $0.9791 \pm 0.0007$ \\
     & -- & Regression only & \multicolumn{1}{c}{--} & \multicolumn{1}{c}{--} & $1.40 \pm 0.02$ & $0.9801 \pm 0.0003$ \\
     & 4-class & Classification only & $85.87 \pm 0.64$ & $54.19 \pm 0.51$ & \multicolumn{1}{c}{--} & \multicolumn{1}{c}{--} \\
     & 4-class & Multi-task & $88.38 \pm 2.40$ & $59.13 \pm 0.67$ & $1.43 \pm 0.03$ & $0.9790 \pm 0.0008$ \\
    \midrule
    Temporal pseudo-RGB & Binary & Classification only & $93.26 \pm 0.44$ & \multicolumn{1}{c}{--} & \multicolumn{1}{c}{--} & \multicolumn{1}{c}{--} \\
     & Binary & Multi-task & $93.24 \pm 0.28$ & \multicolumn{1}{c}{--} & $1.20 \pm 0.04$ & $0.9852 \pm 0.0009$ \\
     & -- & Regression only & \multicolumn{1}{c}{--} & \multicolumn{1}{c}{--} & $1.25 \pm 0.06$ & $0.9784 \pm 0.0105$ \\
     & 4-class & Classification only & $92.60 \pm 0.35$ & $66.00 \pm 0.44$ & \multicolumn{1}{c}{--} & \multicolumn{1}{c}{--} \\
     & 4-class & Multi-task & $93.26 \pm 0.11$ & $66.81 \pm 0.56$ & $1.31 \pm 0.05$ & $0.9750 \pm 0.0090$ \\
    \bottomrule
  \end{tabular*}
\end{table*}

\begin{table*}[t]
  \centering
  \caption{Sensitivity of the result to the temporal span. The three offsets were fixed before any model was trained; this is a sensitivity analysis on the held-out partition, not a selection procedure. Accuracies are percentages. Values are mean $\pm$ SD across five random initializations.}
  \label{tab:2}
  \footnotesize
  \setlength{\tabcolsep}{5pt}
  \renewcommand{\arraystretch}{1.15}
  \begin{tabular*}{\linewidth}{@{\extracolsep{\fill}}lcrrr@{}}
    \toprule
    \textbf{Offset configuration} & \textbf{Span} & \textbf{Binary accuracy} & \textbf{4-class accuracy} & \textbf{MAE (h)} \\
    \midrule
    Single-frame $[0,\,0,\,0]$ & 0 h & $88.38 \pm 2.40$ & $59.13 \pm 0.67$ & $1.43 \pm 0.03$ \\
    Short $[-3,\,-1.5,\,0]$ & 3 h & $91.94 \pm 0.62$ & $63.36 \pm 0.80$ & $1.39 \pm 0.12$ \\
    Standard $[-6,\,-3,\,0]$ & 6 h & $93.26 \pm 0.11$ & $66.81 \pm 0.56$ & $1.31 \pm 0.05$ \\
    Long $[-12,\,-6,\,0]$ & 12 h & $93.01 \pm 0.23$ & $64.30 \pm 0.72$ & $1.35 \pm 0.09$ \\
    \bottomrule
  \end{tabular*}
\end{table*}

\subsection{Comparison against temporal baselines}
\label{sec:baselines}

Encoding the three frames into the input channels is competitive with purpose-built temporal architectures at a fraction of their cost, and on this data it is numerically the most accurate of them (Table~\ref{tab:3}). These are point comparisons under the fixed split, summarized over five seeds. Each was tested at the level of the well, over the 12 wells of the held-out row, using the same paired sign-flip test as the headline comparison, evaluated here by complete enumeration of all $2^{12} = 4096$ sign assignments, with a Holm correction across the five tests. Only the comparison with the single-frame model remains statistically significant once that correction is applied: $+4.88$ percentage points, with all 12 wells favoring the encoding (exact $p = 0.0005$, the smallest value 12 wells can attain; Holm-adjusted $p = 0.0024$). The differences from optical flow, the recurrent model, the 3D network and the published comparator are not resolvable with 12 wells, so the ordering among the learned models should be interpreted descriptively rather than as evidence of pairwise performance differences. A ResNet50 with an LSTM over the same three frames reaches $91.20 \pm 1.44$ and a 3D convolutional network reaches $91.86 \pm 0.57$, against $93.26 \pm 0.11$ for the proposed encoding, which is also the least variable of the three. The single-frame reimplementation of Petkidis et al.\ reaches $90.64 \pm 0.18$ \cite{ref-petkidis2024natcomm}. Trained instead under our own frozen protocol from ImageNet initialization, the same EfficientNet-B0 architecture reaches $89.30 \pm 0.17$ (Section~\ref{sec:backbones}); the published training recipe is therefore the stronger of the two on this data. The cost difference is large: one training run takes $2.0$\,h for the proposed encoding against $6.2$\,h for the recurrent model and $9.6$\,h for the 3D network, on identical hardware and the same 30-epoch schedule.

The temporal baselines differ in the amount of input information available to them. The recurrent model, the 3D network and the proposed encoding are given exactly the same three time-offset frames, so among those three the comparison isolates how the temporal information is used rather than how much of it is supplied. The single-frame model receives only the current frame of the same sample, replicated across the three channels; that difference in input is what the paper compares. The optical-flow baseline is the exception: it accumulates flow over about 13 consecutive frames and therefore receives more input than the other temporal models and achieves $89.64 \pm 0.27$. The non-deep baselines separate into two groups, and accuracy alone does not distinguish them on a $3{:}1$ task, which is why Supplementary Table~S1 reports balanced accuracy, sensitivity and specificity beside it. Four of the six predict the infected class for a far larger share of samples than the $75\%$ prevalence, so their specificities fall to between $0.02$ and $23.4\%$ and their balanced accuracies to between $49.4$ and $55.4\%$, at or barely above chance. Three of those four are almost completely degenerate, predicting infected for $98.5$ to $99.6\%$ of samples; the confluence cut-off is also biased towards the majority label but less extremely, at $84.7\%$, which is why it recovers some specificity and has the highest balanced accuracy of the four. Their accuracies sit at or below the majority-class rate of $75.00$ for that reason: a classifier that answers with the majority label almost always scores near the majority rate, and below it once it also errs on that majority. The frame-difference and flow logistic regressions are different in kind --- balanced accuracies of $78.48$ and $72.62$ and specificities of $65.45$ and $53.63$ --- so those two do carry class information, and they still reach only $85.00$ and $82.11$, below every convolutional model in Table~\ref{tab:3} on binary accuracy. These results suggest that the learned representations capture discriminative information not retained by the hand-crafted temporal summaries evaluated here.

\begin{table*}[t]
  \centering
  \caption{Comparison against temporal baselines, ordered by binary accuracy within each block. Training time is reported because a purpose-built temporal architecture costs several times more than encoding the same frames into the input channels. Accuracies are percentages; dashes denote outputs these baselines do not produce, and training times are wall-clock hours for one 30-epoch run on one NVIDIA A100. Learned-model values are mean $\pm$ SD across five random initializations; the majority-class reference and the non-deep baselines are deterministic given the split and were run once.}
  \label{tab:3}
  \footnotesize
  \setlength{\tabcolsep}{5pt}
  \renewcommand{\arraystretch}{1.15}
  \begin{tabular*}{\linewidth}{@{\extracolsep{\fill}}>{\raggedright\arraybackslash}p{0.30\linewidth}rrrr@{}}
    \toprule
    \textbf{Method} & \textbf{Binary accuracy} & \textbf{4-class accuracy} & \textbf{MAE (h)} & \textbf{Training time (h)} \\
    \midrule
    Majority class & $75.00$\phantom{$\pm 0.00$} & \multicolumn{1}{c}{--} & \multicolumn{1}{c}{--} & \multicolumn{1}{c}{--} \\
    \midrule
    \multicolumn{5}{l}{\itshape non-deep baselines} \\
    Confluence --- single cut-off & $71.39$\phantom{$\pm 0.00$} & \multicolumn{1}{c}{--} & \multicolumn{1}{c}{--} & $<0.1$ \\
    Frame differencing --- single cut-off & $74.06$\phantom{$\pm 0.00$} & \multicolumn{1}{c}{--} & \multicolumn{1}{c}{--} & $<0.1$ \\
    Optical flow --- single cut-off & $74.38$\phantom{$\pm 0.00$} & \multicolumn{1}{c}{--} & \multicolumn{1}{c}{--} & $<0.1$ \\
    Confluence --- logistic regression & $74.90$\phantom{$\pm 0.00$} & \multicolumn{1}{c}{--} & \multicolumn{1}{c}{--} & $<0.1$ \\
    Optical flow --- logistic regression & $82.11$\phantom{$\pm 0.00$} & \multicolumn{1}{c}{--} & \multicolumn{1}{c}{--} & $<0.1$ \\
    Frame differencing --- logistic regression & $85.00$\phantom{$\pm 0.00$} & \multicolumn{1}{c}{--} & \multicolumn{1}{c}{--} & $<0.1$ \\
    \midrule
    \multicolumn{5}{l}{\itshape learned models} \\
    Single-frame ResNet50 & $88.38 \pm 2.40$ & $59.13 \pm 0.67$ & $1.43 \pm 0.03$ & $2.1$ \\
    Optical flow (Farneb\"{a}ck) $\rightarrow$ ResNet50 & $89.64 \pm 0.27$ & $57.13 \pm 0.20$ & $1.67 \pm 0.01$ & $2.0$ \\
    Petkidis et al. 2024 (EfficientNet-B0) & $90.64 \pm 0.18$ & $55.71 \pm 1.49$ & $1.64 \pm 0.06$ & $1.8$ \\
    ResNet50 + LSTM & $91.20 \pm 1.44$ & $64.86 \pm 4.44$ & $1.35 \pm 0.25$ & $6.2$ \\
    3D CNN (\texttt{r3d\_18}) & $91.86 \pm 0.57$ & $63.57 \pm 0.82$ & $1.04 \pm 0.04$ & $9.6$ \\
    Temporal pseudo-RGB (this work) & $93.26 \pm 0.11$ & $66.81 \pm 0.56$ & $1.31 \pm 0.05$ & $2.0$ \\
    \bottomrule
  \end{tabular*}
\end{table*}

\subsection{Well-level statistical analysis and balanced metrics}
\label{sec:stats}

The unit of analysis is the well and the study contains $36$ of them (Section~\ref{sec:protocol}). Each well is scored by the cross-validation fold in which it was held out, so all $36$ contribute exactly once and no well is evaluated by a model that saw it in training.

\begin{figure*}[t]
  \centering
  \includegraphics[width=\linewidth]{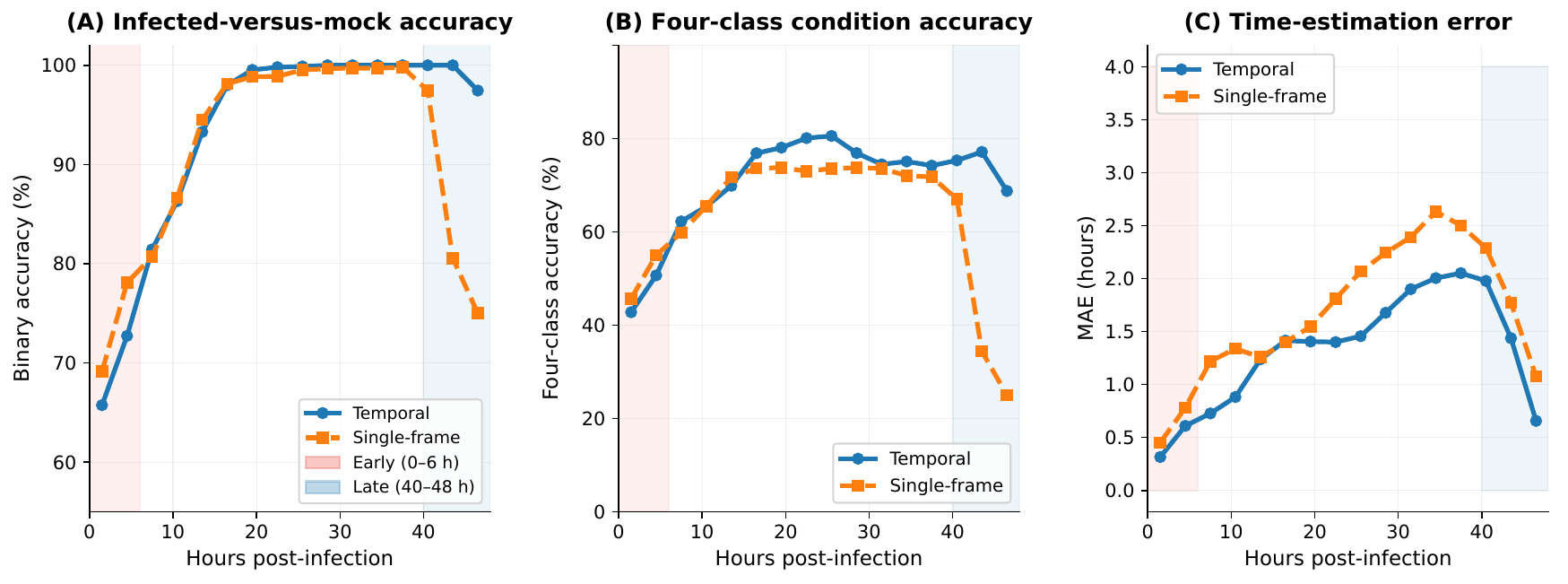}
  \caption{Temporal encoding sustains infected-versus-mock detection across the 48-hour infection timeline. Temporal multi-task model against its single-frame counterpart, in 3-hour bins, from a single representative run on the fixed row split; the corresponding whole-timeline and late-phase values under the leave-one-row-out protocol of the headline comparison are in Table~\ref{tab:4}.
(A) Infected-versus-mock accuracy: the temporal model (blue) holds above 97\% at 40--48\,h while the single-frame model (orange, dashed) falls to roughly 75\%. Shading marks the early (0--6\,h, red) and late (40--48\,h, blue) phases.
(B) Four-class condition accuracy over time, shown for context.
(C) Time post-infection MAE over time. The temporal model has the lower error over most of the time course, by the largest absolute margin at 30--37\,h. These descriptive temporal-versus-single-frame differences were not tested at the well level.}
  \label{fig:2}
\end{figure*}

Across the whole time course, the temporal encoding is more accurate than the single-frame model by $+4.23$ percentage points (cluster-bootstrap 95\% CI $[+2.51, +6.05]$; paired sign-flip permutation $p = 3 \times 10^{-5}$; Wilcoxon signed-rank $p = 3.1 \times 10^{-5}$; Table~\ref{tab:4}), with $31$ of $36$ wells favoring the temporal model, and the paired difference positive within each of the five seeds at both endpoints. Restricted to the late-phase endpoint --- the final eight hours of the time course --- the difference is $+23.87$ percentage points (95\% CI $[+19.16, +28.85]$; sign-flip $p_{\mathrm{MC}} = 1.0 \times 10^{-5}$, the resolution of the reference distribution, none of the $10^5$ sampled sign assignments being as or more extreme than the observed difference; Wilcoxon $p = 2.9 \times 10^{-11}$), and every one of the $36$ wells favors the temporal model. Figure~\ref{fig:2} illustrates, for a single training run on the fixed row split, how the two models separate across the time course in 3-hour bins, with the separation in the final third of the experiment.

Because the 36 wells sit inside only three runs, the tests above are conditional on those runs and cannot determine whether the effect is driven disproportionately by one of them. The observed effect was not confined to a single experimental run: taken run by run, the whole-timeline difference is $+5.33$, $+4.34$ and $+3.04$ percentage points, and the late-phase difference is $+29.48$, $+22.49$ and $+19.64$ percentage points. The direction is the same in all three runs at both endpoints, and at the late-phase endpoint all 12 wells favor the temporal model within every run. The magnitude varies by about 50\% across runs, which is the between-run variability that a well-level test does not itself quantify.

\begin{table*}[t]
  \centering
  \caption{Paired comparison of the temporal encoding against the single-frame model, with the well as the unit of analysis. Each well is scored by the cross-validation fold that held it out, so all 36 wells contribute exactly once. Accuracies are percentages, given to three decimals so that they subtract to the difference beside them; the text rounds to two.}
  \label{tab:4}
  \footnotesize
  \setlength{\tabcolsep}{5pt}
  \renewcommand{\arraystretch}{1.15}
  \begin{tabular*}{\linewidth}{@{\extracolsep{\fill}}lrrrcrr@{}}
    \toprule
    \textbf{Endpoint} & \textbf{Temporal} & \textbf{Single-frame} & \textbf{$\Delta$ (pp)} & \textbf{95\% CI (pp)} & \textbf{p (sign-flip)} & \textbf{p (Wilcoxon)} \\
    \midrule
    Whole time course & $91.619$ & $87.385$ & $+4.234$ & $[+2.51, +6.05]$ & $3{\times}10^{-5}$ & $3.05{\times}10^{-5}$ \\
    Last 8 h (late-phase endpoint) & $99.427$ & $75.556$ & $+23.871$ & $[+19.16, +28.85]$ & $1.0{\times}10^{-5}$ & $2.91{\times}10^{-11}$ \\
    \bottomrule
  \end{tabular*}
\end{table*}

\label{sec:balanced}
\emph{Balanced binary metrics.} 

The four conditions comprise three infected and one mock, so the infected-versus-mock task carries a $3{:}1$ class ratio and the majority-class rate is $75\%$. Reporting accuracy alone would be insufficient, and the balanced metrics show that the advantage is not an artifact of the imbalance (Table~\ref{tab:5}). Over the whole time course, the temporal model reaches balanced accuracy $88.85 \pm 0.25$ against $83.06 \pm 0.13$ for the single-frame model, F1 $94.41 \pm 0.07$ against $91.55 \pm 1.71$, AUROC $0.9671 \pm 0.0017$ against $0.9325 \pm 0.0038$, and AUPRC $0.9876 \pm 0.0013$ against $0.9746 \pm 0.0023$. The gain is larger in specificity ($83.31 \pm 0.56$ versus $74.42 \pm 4.04$) than in sensitivity ($94.40 \pm 0.13$ versus $91.70 \pm 4.23$): the single-frame model's errors are predominantly mock fields called infected, which is the failure mode the temporal input is intended to remove.

The single-frame arm's one-sided rates are also the least reproducible numbers in the study, and the pattern of that instability is informative. In the late phase its sensitivity and specificity vary across seeds with standard deviations of $23.6$ and $24.1$ percentage points, while its balanced accuracy varies by $0.44$ and its AUROC by $0.0009$. A model whose separability is stable but whose two one-sided rates trade against each other is one that lands on a different decision threshold at each initialization rather than one that has learned something different. For the single-frame model, therefore, balanced accuracy and AUROC provide the more stable late-phase summary measures, whereas sensitivity or specificity considered individually is highly seed-dependent in this setting. The temporal model shows nothing comparable: every one of its late-phase metrics sits within $0.5$ percentage points across seeds.

\begin{table}[t]
  \centering
  \caption{Balanced binary metrics for the infected-versus-mock readout, computed on the pooled out-of-fold predictions across all 36 wells, separately for each of the five seeds. The positive class is infected, and the class ratio is 3:1. Values are mean $\pm$ SD across five random initializations.}
  \label{tab:5}
  \footnotesize
  \setlength{\tabcolsep}{5pt}
  \renewcommand{\arraystretch}{1.15}
  \begin{tabular}{lrr}
    \toprule
    \textbf{Metric} & \textbf{Temporal pseudo-RGB} & \textbf{Single-frame} \\
    \midrule
    Balanced accuracy & $88.85 \pm 0.25$ & $83.06 \pm 0.13$ \\
    Sensitivity & $94.40 \pm 0.13$ & $91.70 \pm 4.23$ \\
    Specificity & $83.31 \pm 0.56$ & $74.42 \pm 4.04$ \\
    Precision & $94.42 \pm 0.17$ & $91.52 \pm 0.92$ \\
    F1 & $94.41 \pm 0.07$ & $91.55 \pm 1.71$ \\
    AUROC (0--1) & $0.9671 \pm 0.0017$ & $0.9325 \pm 0.0038$ \\
    AUPRC (0--1) & $0.9876 \pm 0.0013$ & $0.9746 \pm 0.0023$ \\
    \bottomrule
  \end{tabular}
\end{table}

\subsection{Performance across network backbones}
\label{sec:backbones}

The encoding requires no architectural modification for backbones that accept a three-channel input. Across three convolutional architectures the gain is consistent: $+4.88$ percentage points for ResNet50, $+2.94$ for EfficientNet-B0 and $+3.21$ for ConvNeXt-Tiny (Table~\ref{tab:6}).

In contrast, ViT-B/16 showed no improvement with temporal encoding. Its temporal arm reaches $88.40 \pm 2.26$ against $88.74 \pm 0.10$ for its single-frame arm, a difference of $-0.34$ percentage points, and it is the least stable configuration in the study --- its standard deviation across seeds is more than 20 times that of the corresponding single-frame arm. Under the present protocol the observed advantage is therefore confined to the three convolutional backbones tested. This comparison does not isolate an architectural cause: the size of the seed-to-seed spread points at optimization rather than representation, and the transformer arm also differs from its pretraining regime in input resolution and hence in the number of patches, which this study does not disentangle.

\begin{table}[t]
  \centering
  \caption{The temporal encoding applied to four ImageNet-pretrained backbones, each trained under the same protocol with and without the encoding. Accuracies are percentages. Values are mean $\pm$ SD across five random initializations.}
  \label{tab:6}
  \footnotesize
  \setlength{\tabcolsep}{5pt}
  \renewcommand{\arraystretch}{1.15}
  \begin{tabular}{lrrr}
    \toprule
    \textbf{Backbone} & \textbf{Temporal} & \textbf{Single-frame} & \textbf{$\Delta$ (pp)} \\
    \midrule
    ResNet50 & $93.26 \pm 0.11$ & $88.38 \pm 2.40$ & $+4.88$ \\
    EfficientNet-B0 & $92.24 \pm 0.38$ & $89.30 \pm 0.17$ & $+2.94$ \\
    ConvNeXt-Tiny & $92.18 \pm 0.26$ & $88.97 \pm 2.23$ & $+3.21$ \\
    ViT-B/16 & $88.40 \pm 2.26$ & $88.74 \pm 0.10$ & $-0.34$ \\
    \bottomrule
  \end{tabular}
\end{table}

\subsection{Positional and normalization controls}
\label{sec:posresults}

Plate-position effects, such as edge evaporation, thermal or stage gradients and illumination non-uniformity, could in principle be confounded with infection, since every plate used the same condition-to-column layout. Within each well, where condition, run and fold are fixed and only physical position varies, we found no detectable edge-to-interior accuracy difference. The paired edge-minus-interior difference is $+0.02$ percentage points over the whole time course for the temporal model (95\% CI $[-0.74, +0.75]$) and $-0.08$ for the single-frame model (CI $[-0.67, +0.49]$); in the late phase it is $-0.09$ (CI $[-0.31, +0.12]$) and $-0.04$ (CI $[-0.18, +0.10]$). Restricted to the nine mock wells, where no infection signal exists and any positional structure must be positional, the difference is $-1.20$ percentage points with a wider interval (CI $[-3.38, +0.83]$) that also contains zero. Every interval in this set straddles zero, and the widest of them, $\pm3.4$ percentage points, is the resolution this design supports.

\emph{Normalization sensitivity.} 

Per-frame min--max normalization is sensitive to single-frame artifacts. We therefore evaluated the sensitivity of the result to the normalization procedure. Both models were retrained from scratch under 1st--99th percentile normalization, which bounds the influence of single-pixel extrema (Table~\ref{tab:norm}). The temporal advantage is present under both pipelines: $+5.66$ percentage points under the percentile variant and $+4.88$ under the reported pipeline. We retain per-frame min--max as the reported pipeline, since it is the convention in this literature and the comparison shows the conclusion does not rest on the choice.
\begin{table*}[t]
  \centering
  \caption{Sensitivity of the temporal advantage to the intensity normalization. Both models were retrained from scratch under each pipeline, because re-evaluating existing models on renormalized images would measure robustness to an inference-time shift rather than the consequence of the normalization choice. Accuracies are percentages. Values are mean $\pm$ SD across five random initializations.}
  \label{tab:norm}
  \footnotesize
  \setlength{\tabcolsep}{5pt}
  \renewcommand{\arraystretch}{1.15}
  \begin{tabular*}{\linewidth}{@{\extracolsep{\fill}}>{\raggedright\arraybackslash}p{0.30\linewidth}rrr@{}}
    \toprule
    \textbf{Normalization} & \textbf{Temporal} & \textbf{Single-frame} & \textbf{$\Delta$ (pp)} \\
    \midrule
    Per-frame min--max (reported pipeline) & $93.26 \pm 0.11$ & $88.38 \pm 2.40$ & $+4.88$ \\
    1st--99th percentile & $93.59 \pm 0.28$ & $87.93 \pm 2.48$ & $+5.66$ \\
    \bottomrule
  \end{tabular*}
\end{table*}

\subsection{Cross-validated performance by fold, dose and phase}
\label{sec:cv}

Because the study contains only three plates and one row of each is held out, a single split leaves the result dependent on which row that is. Repeating the protocol as leave-one-row-out cross-validation removes that dependence: each fold holds out one row across all three plates, every well is held out exactly once, and the train/test proportion is constant (Table~\ref{tab:s1}). The temporal model reaches $87.69 \pm 0.35$, $93.89 \pm 0.12$ and $93.26 \pm 0.11$ on folds A, B and C, against $84.81 \pm 2.06$, $88.92 \pm 3.00$ and $88.38 \pm 2.40$ for the single-frame model. The advantage is present in every fold, ranging from $+2.88$ to $+4.97$ percentage points. Averaging each well's accuracy over the five seeds and then over the 36 out-of-fold wells gives $91.62$ against $87.38$; this is the quantity the paired tests of Table~\ref{tab:4} are computed on, and it is the figure quoted in the abstract.

Two observations are relevant to the interpretation of these results. Absolute accuracy varies more across folds than across seeds --- the temporal model spans $87.7$ to $93.9$ depending on which row is held out, against a seed-to-seed spread of about $0.4$ percentage points within a fold --- so the choice of split, not optimization noise, is the dominant source of uncertainty in a single-split estimate. And the single-frame model is the more variable of the two arms in every fold, by roughly an order of magnitude.

\begin{table}[t]
  \centering
  \caption{Leave-one-row-out cross-validation. Each fold holds out one row across all three plates, so every well is held out exactly once and the train/test proportion is constant. Accuracies are percentages; the pooled row is given to three decimals so that it subtracts to the difference beside it, and the text rounds to two. Values are mean $\pm$ SD across five random initializations.}
  \label{tab:s1}
  \footnotesize
  \setlength{\tabcolsep}{5pt}
  \renewcommand{\arraystretch}{1.15}
  \begin{tabular}{lrrc}
    \toprule
    \textbf{Fold} & \textbf{Temporal} & \textbf{Single-frame} & \textbf{$\Delta$ (pp)} \\
    \midrule
    Fold A & $87.69 \pm 0.35$ & $84.81 \pm 2.06$ & $+2.88$ \\
    Fold B & $93.89 \pm 0.12$ & $88.92 \pm 3.00$ & $+4.97$ \\
    Fold C & $93.26 \pm 0.11$ & $88.38 \pm 2.40$ & $+4.88$ \\
    \midrule
    Pooled out-of-fold & $91.619$\phantom{$\pm 0.00$} & $87.385$\phantom{$\pm 0.00$} & $+4.234$ \\
    \bottomrule
  \end{tabular}
\end{table}

\label{sec:dose}
\emph{Reliability by dose and by phase.} 

The same 36 out-of-fold wells allow detectability to be read against the two experimentally relevant variables for screening applications: the multiplicity of infection and the point in the time course (Table~\ref{tab:perwell}; the binned time course of both models is shown in Figure~\ref{fig:2}, and per-condition AUROC in 3-hour bins in Figure~S9). Each cell of Table~\ref{tab:perwell} summarizes nine wells --- three technical replicates on each of three plates --- and the bracketed ranges are included to show between-well variability rather than the condition means alone. These are descriptive summaries: the paired tests of Table~\ref{tab:4} are computed over all 36 wells together, and no within-condition significance testing was performed, because each condition comprises only nine wells nested within three runs.

Over the whole time course, the temporal model is the more accurate of the two at every condition. The late phase separates them sharply, and it does so unevenly across conditions. On the three infected conditions the single-frame model still reaches roughly $80$ in the final eight hours while the temporal model reaches $99.4$ or above; on Mock it falls to $59.8$, with its worst well at $30.8$, against $98.5$ for the temporal model. The late-phase failure is therefore concentrated on the condition that carries no infection signal at all, which is the well-level counterpart of the specificity gap in Table~\ref{tab:5} and of the upward drift of the single-frame model's Mock confidence trajectory (Figure~S1). An endpoint plaque assay of the culture supernatants was positive in all three infected conditions and negative in mock (Figure~S7), confirming that the infections were productive.

The bands also locate the instability that motivates our treatment of the low-multiplicity results. Over the whole time course, MOI~0.1 has the widest spread of the three infected conditions, $74.1$ to $100.0$ across nine wells. That spread persists at this level of replication rather than narrowing, and the low-multiplicity results are reported as exploratory for that reason (Section~\ref{sec:limitations}). Every well is listed individually in Supplementary Table~S2.

As an additional within-study robustness check across experimental runs, we held out one entire run at a time, so that the model was evaluated on a plate not represented in training. Under these folds the temporal encoding is more accurate than the single-frame model by $+5.8$ percentage points over the whole time course and by $+28.5$ points at the late-phase endpoint, with 35 of 36 and all 36 wells favoring it respectively, and the same direction in each of the three folds (Table~\ref{tab:lororun}).

\begin{table*}[t]
  \centering
  \caption{Detection reliability by infection condition and by phase of the time course. Each cell is the mean over that condition's nine wells of the well's accuracy, with the range across those nine wells in brackets.}
  \label{tab:perwell}
  \footnotesize
  \setlength{\tabcolsep}{5pt}
  \renewcommand{\arraystretch}{1.15}
  \begin{tabular*}{\linewidth}{@{\extracolsep{\fill}}lc>{\centering\arraybackslash}p{0.155\linewidth}>{\centering\arraybackslash}p{0.155\linewidth}>{\centering\arraybackslash}p{0.155\linewidth}>{\centering\arraybackslash}p{0.155\linewidth}@{}}
    \toprule
     &  & \multicolumn{2}{c}{\textbf{Whole time course}} & \multicolumn{2}{c}{\textbf{Late phase 40--48\,h}} \\
    \cmidrule(lr){3-4}\cmidrule(lr){5-6}
    \textbf{Condition} & \textbf{Wells} & \textbf{Temporal} & \textbf{Single-frame} & \textbf{Temporal} & \textbf{Single-frame} \\
    \midrule
    MOI 5 & $9$ & 92.1 [79.3--100.0] & 89.7 [75.0--97.9] & 99.4 [96.5--100.0] & 80.4 [53.6--91.9] \\
    MOI 1 & $9$ & 97.4 [89.7--100.0] & 95.0 [87.6--98.6] & 99.9 [99.4--100.0] & 80.8 [55.0--91.6] \\
    MOI 0.1 & $9$ & 93.6 [74.1--100.0] & 90.4 [74.5--98.1] & 100.0 [100.0--100.0] & 81.1 [58.2--91.2] \\
    Mock & $9$ & 83.3 [55.9--98.0] & 74.4 [38.9--92.8] & 98.5 [88.5--100.0] & 59.8 [30.8--82.3] \\
    \bottomrule
  \end{tabular*}
\end{table*}

\begin{table}[t]
  \centering
  \caption{Leave-one-run-out cross-validation. Each fold holds out all twelve wells of one experimental run, so the model is scored on a plate it never saw in training. Accuracies are percentages. Values are mean $\pm$ SD across five random initializations, and differences are computed before rounding. No test of the difference between this protocol and Table~\ref{tab:s1} was performed, and none is claimed.}
  \label{tab:lororun}
  \footnotesize
  \setlength{\tabcolsep}{5pt}
  \renewcommand{\arraystretch}{1.15}
  \begin{tabular}{lrrc}
    \toprule
    \textbf{Held-out run} & \textbf{Temporal} & \textbf{Single-frame} & \textbf{$\Delta$ (pp)} \\
    \midrule
    Run 1 & $97.59 \pm 0.18$ & $90.50 \pm 3.23$ & $+7.08 \pm 3.24$ \\
    Run 2 & $87.32 \pm 0.67$ & $82.98 \pm 2.25$ & $+4.34 \pm 1.90$ \\
    Run 3 & $92.01 \pm 0.42$ & $86.07 \pm 1.78$ & $+5.94 \pm 1.80$ \\
    \midrule
    Pooled out-of-fold & $92.31$\phantom{$\pm 0.00$} & $86.52$\phantom{$\pm 0.00$} & $+5.79$\phantom{$\pm 0.00$} \\
    \bottomrule
  \end{tabular}
\end{table}

\section{Discussion}
\subsection{Temporal encoding and the late-phase divergence}

The central observation of this study is that encoding three time-offset frames, rather than the instantaneous appearance alone, was associated with more stable label-free discrimination between infected and mock cultures in this time-lapse brightfield system.
The effect is present across the whole time course ($+4.23$ percentage points, 95\% CI $[+2.51, +6.05]$) but is concentrated at the late-phase endpoint, 40--48\,h, where it reaches $+23.87$ points (95\% CI $[+19.16, +28.85]$) and where every one of the 36 wells favors the temporal model.
This matters for screening and longitudinal imaging applications, which require detection that holds across a whole experiment rather than only during the intermediate window in which CPE is most conspicuous.

A plausible explanation for the late-phase drop in the single-frame model is that floating cells and debris obscure monolayer morphology in conventional brightfield images, reducing the information available in an isolated snapshot.
The predicted infection probability follows the same pattern over time: the single-frame model's Mock trajectory drifts upward late in the experiment while the temporal model's does not (Figure~S1), and the two models' backbone features are organized differently overall (Figure~S4). The balanced metrics are consistent with this reading: the single-frame model's disadvantage is larger in specificity ($74.42 \pm 4.04$ against $83.31 \pm 0.56$) than in sensitivity ($91.70 \pm 4.23$ against $94.40 \pm 0.13$), that is, it fails predominantly by calling mock fields infected.
These analyses identify the late-phase failure but do not determine its visual cause; the Grad-CAM \cite{ref-selvaraju2017gradcam} maps of Figure~S5 are suggestive rather than confirmatory.

Existing deep-learning approaches to CPE detection---including Petkidis et al. \cite{ref-petkidis2024natcomm} on adenovirus and HSV, Andriasyan et al. \cite{ref-andriasyan2021iscience} on adenovirus, Chen et al. \cite{ref-chen2022biomedicines} on influenza, parainfluenza and enterovirus, and AIRVIC \cite{ref-airvic2025}---operate on isolated frames.
Their published accuracies are not directly comparable to ours, because the virus--cell systems, datasets and evaluation protocols differ; we therefore reimplemented the training recipe of Petkidis et al.\ on our data and fixed split, retaining its EfficientNet-B0 architecture, random initialization, optimizer and augmentation (Table~\ref{tab:3}), and separately trained the same architecture under our own frozen protocol from ImageNet initialization (Table~\ref{tab:6}). The first holds the data and the evaluation constant while changing several things at once; the second changes only the architecture.
Under that comparison it reaches $90.64 \pm 0.18$ against $93.26 \pm 0.11$ for the temporal encoding.
The studies reviewed above did not examine late-stage morphological ambiguity, nor characterize when reliable detection becomes available as a function of multiplicity of infection and time post-infection.

The simplicity of pseudo-RGB encoding is a deliberate design choice.
Brightfield images are inherently single-channel; mapping three time-offset frames to the three channels expected by an ImageNet-pretrained CNN reuses color-processing features the network has already learned from natural images, with no architectural change and no purpose-built temporal module.
On this data that choice was cheaper than the two purpose-built alternatives and numerically ahead of them: a CNN--LSTM and a 3D convolutional network over the same three frames reached $91.20 \pm 1.44$ and $91.86 \pm 0.57$ against $93.26 \pm 0.11$, at roughly three and five times the training cost. Tested at the well level over the 12 held-out wells, neither difference is resolvable; these results therefore do not establish superiority over purpose-built temporal architectures. With 36 wells, the present study cannot resolve differences among these temporal modeling approaches.
The advantage was observed across all evaluated offset configurations (Section~\ref{sec:ablation}).

\subsection{Effect of auxiliary time-regression supervision}

The present comparisons do not provide evidence that auxiliary time-regression supervision improves infected-versus-mock classification. With a binary head, classification-only and multi-task training yielded similar mean accuracies across the five initializations ($93.26 \pm 0.44$ against $93.24 \pm 0.28$); with a four-class head the multi-task arm is 0.66 percentage points ahead ($92.60 \pm 0.35$ against $93.26 \pm 0.11$), a contrast that was not tested at the well level and is therefore not interpreted as evidence of improved detection. A single-seed comparison would not separate these arms, because the seed-to-seed variation is of the same order as the difference between them.

For time estimation, however, joint training produced a small reduction in error: MAE $= 1.20 \pm 0.04$\,h against $1.25 \pm 0.06$\,h for a regression-only model on the same input, with $R^2$ $0.9852 \pm 0.0009$ against $0.9784 \pm 0.0105$ --- the direction the general observation about related objectives regularizing a shared representation would predict \cite{ref-caruana1997mtl,ref-ruder2017mtloverview}. The well-level comparison reported in Section~\ref{sec:ablation} resolved that reduction, about 3 minutes on an error of roughly 75 minutes, although it was concentrated in one of the three runs, and it is not on its own a reason to prefer the joint objective. The primary practical advantage is efficiency rather than improved classification accuracy: one forward pass returns both an infected-versus-mock call and an estimate of the time elapsed since inoculation.

An important consideration is the granularity of the ground truth.
Infection labels are defined by experimental well condition rather than by single-cell confirmation within each image field.
This is especially relevant at low MOI, where infected and uninfected cells coexist within a well and morphological change is spatially heterogeneous.
The task studied here is accordingly best interpreted as discrimination between image fields drawn from infected versus mock wells, not as quantification of infection status in individual cells.

\subsection{Backbone dependence of temporal pseudo-RGB encoding}

Applied to three convolutional backbones under an identical protocol, the encoding gave a consistent advantage: $+4.88$ percentage points for ResNet50, $+2.94$ for EfficientNet-B0 and $+3.21$ for ConvNeXt-Tiny.
Applied to ViT-B/16 it gave none ($-0.34$ points), and that configuration was the least stable in the study, with a seed-to-seed standard deviation more than 20 times that of its own single-frame arm.
Compatibility with a three-channel input does not by itself imply that the performance benefit transfers across architecture families.

We do not attribute this to architecture. The temporal ViT arm is the least stable configuration in the study, and its input resolution, data requirements and optimization behavior differ from the convolutional backbones in ways this experiment does not separate; prior work on the encoding tested no transformer \cite{ref-kim2022grayst}.
Under the present data size and training protocol, the ViT-B/16 comparison did not show an improvement from the encoding. Evaluation of a single transformer architecture at one input scale is insufficient to draw broader conclusions about vision transformers.

\subsection{Limitations of the study}
\label{sec:limitations}

The principal limitation is the amount of independent biological material.
The study comprises three experimental runs and 36 wells in total, 12 per run, all from a single commercial cell lot. The wells are nested within the runs, so they are not 36 independent biological units, and the reported intervals and p-values are conditional on these three runs rather than generalizing over runs. The leave-one-run-out analysis of Section~\ref{sec:cv} speaks to part of this: the temporal advantage was still observed when each experimental run in turn was excluded from training. That analysis remains within this study and is not external validation, which was not performed.
The roughly $1.2\times10^{5}$ frames represent dense repeated sampling of those 36 wells rather than $1.2\times10^{5}$ independent observations.
Every interval and p-value for that comparison rests on those 36 wells; the baseline comparisons of Table~\ref{tab:3} are on the fixed split and rest on the 12 wells of the held-out row, fewer still.
The late-phase effect is large and consistent across wells, whereas the present design has limited precision for differences on the order of one percentage point, such as the multi-task comparison reported above.

Second, the experimental scope is narrow in four independent respects, and the conclusions are stated only for it: a single commercially obtained HBMVEC preparation used at passages 4--6 as a monoculture without astrocytes or pericytes; a single virus strain, the attenuated vaccine strain TC-83, whose CPE kinetics need not match those of virulent VEEV or of other alphaviruses; a single imaging platform and acquisition mode; and a single laboratory. No external validation on independently generated data was performed.

Third, residual positional confounding cannot be excluded by design. Because infection condition was mapped to the same plate column in all three runs, no well-level stratification can separate plate-level positional effects from condition effects. Holding out a plate row or an entire run does not change that mapping, so neither cross-validation scheme separates them. The within-well edge-versus-interior comparison, in which condition, run and fold are held constant and only physical position varies, detected no spatial accuracy gradient in either model at either endpoint, and none in the mock wells alone (Section~\ref{sec:posresults}). That control bounds a within-well gradient rather than a plate-level offset that would shift a whole well together, which the available analysis cannot exclude. Randomized or counterbalanced plate layouts remain the appropriate remedy for future work.

Fourth, detection at the lowest multiplicity (MOI~0.1) is markedly more variable across experimental runs than at higher multiplicities, consistent with only a small fraction of cells being infected initially and with heterogeneous, delayed CPE accumulation.
The MOI~0.1 results should therefore be regarded as exploratory rather than as an established operating point.

Fifth, per-frame min--max normalization is sensitive to single-frame artifacts, in the specific way set out in Section~\ref{sec:qc} and the encoding description: an artifact confined to one offset produces a global color cast rather than a local change, a failure mode that cannot arise in the single-frame configuration.
Relatedly, the per-well confluence measurements used for quality control are intrinsically variable, because the segmentation mask must accept cells both as an adherent monolayer and as they round up and detach; this is a limitation of that measurement rather than of the imaging.

Sixth, the encoding requires sequential imaging and introduces a 6\,h latency before a complete three-channel input exists; clamp padding fills the earliest frames and reduces the effective temporal information there.
The offsets ($-$6, $-$3, 0\,h) were not tuned and are unlikely to be optimal across virus--cell systems with different kinetics; adaptive offset selection is a natural direction for future work.

Finally, infection labels are assigned at the well level from the experimental condition; individual cells within infected wells may remain uninfected, particularly at low MOI, and orthogonal single-cell validation such as immunofluorescence was not performed.

\section{Conclusions}

Building on a previously described temporal channel-stacking representation---three time-offset brightfield frames written into the red, green and blue channels of a single image---we evaluated temporal pseudo-RGB encoding for label-free detection of VEEV TC-83 infection in human brain microvascular endothelial cells, in a regime where the offsets span hours and the change of interest is slow and non-rigid rather than rigid motion.

In that regime the encoding was more accurate than an otherwise identical single-frame model by $+4.23$ percentage points over the whole 48-hour time course and by $+23.87$ points over the final eight hours, the late-phase endpoint, under leave-one-row-out cross-validation with the well as the unit of analysis, with the same sign in each of the three runs. The advantage also persisted when each experimental run was withheld in turn.
Among the temporal models evaluated it achieved the highest mean accuracy at substantially lower training cost than the CNN--LSTM and 3D convolutional alternatives given the same three frames, although differences among these temporal models were not statistically resolved.
The gain was observed across three convolutional backbones; the ViT-B/16 comparison showed no corresponding improvement, which limits the claim to convolutional architectures under this data size and protocol.
The auxiliary time-regression objective improves the time estimate and yields two readouts from one forward pass, but the present comparisons do not provide evidence of improved detection accuracy.

The principal finding is an observed late-phase divergence between the two procedures: single-frame classification degrades specifically in the final eight hours of the time course, and temporal encoding showed substantially less of that degradation, with little additional computational cost.
These results were obtained in one cell preparation, one virus strain, one imaging platform and one laboratory, on 36 wells, and should be read within that scope.
Whether the encoding transfers to other virus--cell systems, to virulent strains, and to independently generated data remains open.

\printcredits

\section*{Declaration of competing interest}
The authors declare no competing interests.

\section*{Declaration of generative AI and AI-assisted technologies in the writing process}
During the preparation of this work, the authors used Claude (Anthropic) and Codex (OpenAI) to assist with manuscript formatting and language editing. No image representing primary experimental data was altered by these tools, and no experimental result was generated, selected, or modified by them. After using these tools, the authors reviewed and edited the content as needed and take full responsibility for the content of the publication.

\section*{Funding}
This research did not receive any specific grant from funding agencies in the public, commercial, or not-for-profit sectors.

\section*{Data availability}
\begin{sloppypar}
Requests for further information, resources, and data access should be directed to Zhengjie Zhu (\href{mailto:Zhengjie.Zhu@wfusm.edu}{Zhengjie.Zhu@wfusm.edu}).
\end{sloppypar}
\begin{itemize}
  \item The microscopy image dataset reported in this paper is not publicly available because of confidentiality restrictions associated with the underlying study materials. Requests for access to non-public data should be directed to the corresponding author and will be considered subject to institutional and data-use restrictions.
  \item All original code has been deposited at GitHub (\url{https://github.com/CAIR-LAB-WFUSM/cell_tempo}) and is publicly available as of the date of publication.
  \item The out-of-fold accuracy of every well in the study, which is the input to every significance test reported here, is given in full in Supplementary Table~S2.
\end{itemize}

\section*{Acknowledgements}
The authors thank the members of the Gurcan and Narayanan laboratories for helpful discussions and technical support.

\bibliographystyle{elsarticle-num}
\bibliography{references}

\end{document}